\documentclass[10pt,twocolumn,twoside]{IEEEtran}

\usepackage{graphicx}
\usepackage{comment}
\usepackage[misc,geometry]{ifsym} 
\usepackage{color}
\usepackage{hyperref}
\usepackage[bottom]{footmisc}
\usepackage{supertabular}
\usepackage{afterpage}
\usepackage{url}
\usepackage{pifont}
\usepackage{multicol}
\usepackage{multirow}
\usepackage[table,xcdraw, dvipsnames]{xcolor}
\usepackage{adjustbox}
\usepackage{tabularx}
\usepackage{array}
\usepackage{amsmath}
\usepackage{amsfonts}
\usepackage{subcaption}
\usepackage{cite}
\usepackage[square,sort,comma,numbers]{natbib}
\usepackage[linesnumbered,ruled,vlined]{algorithm2e}
\SetKwInput{KwInput}{Input}
\SetKwInput{KwOutput}{Output}
\definecolor{gray}{cmyk}{0,0,0,0.70}
\newcommand{\blue}[1]{{\color{blue}\textbf{#1}}}
\newcommand{\green}[1]{{\color{Green}\textbf{#1}}}
\newcommand{\red}[1]{{\color{red}\textbf{#1}}}
\newcolumntype{Y}{>{\centering\arraybackslash}X}

\definecolor{orcidlogo}{rgb}{0.37,0.48,0.13}
\definecolor{unilogo}{rgb}{0.16, 0.26, 0.58}
\definecolor{maillogo}{rgb}{0.58, 0.16, 0.26}
\definecolor{darkblue}{rgb}{0.0,0.0,0.0}
\hypersetup{colorlinks,breaklinks,
            linkcolor=darkblue,urlcolor=darkblue,
            anchorcolor=darkblue,citecolor=darkblue}
\usepackage[normalem]{ulem}

\title{FLIM-based Salient Object Detection Networks with Adaptive Decoders}

\author{
\IEEEauthorblockN{Gilson Junior Soares\IEEEauthorrefmark{1},~Matheus Abrantes Cerqueira\IEEEauthorrefmark{1},~Jancarlo F. Gomes\IEEEauthorrefmark{1},~Laurent Najman\IEEEauthorrefmark{2},~Silvio Jamil F. Guimarães\IEEEauthorrefmark{3} and Alexandre Xavier Falc\~{a}o\IEEEauthorrefmark{1}}

\IEEEauthorblockA{\IEEEauthorrefmark{1}Institute of Computing, State University of Campinas, Campinas, 13083-872, São Paulo, Brazil}

\IEEEauthorblockA{\IEEEauthorrefmark{2}Univ Gustave Eiffel, CNRS, LIGM}

\IEEEauthorblockA{\IEEEauthorrefmark{3}~Pontificial Catholic University of Minas Gerais}

\IEEEauthorblockA{\{gilson.soares, matheus.cerqueira\}@students.ic.unicamp.br\\laurent.najman@esiee.fr sjamil@pucminas.br\\ \{jgomes, afalcao\}@unicamp.br}
}

\begin{document}

\maketitle

\begin{abstract}
Salient Object Detection (SOD) methods can locate objects that stand out in an image, assign higher values to their pixels in a saliency map, and binarize the map outputting a predicted segmentation mask. A recent tendency is to investigate pre-trained lightweight models rather than deep neural networks in SOD tasks, coping with applications under limited computational resources. In this context, we have investigated lightweight networks using a methodology named \textit{Feature Learning from Image Markers} (FLIM), which assumes that the encoder's kernels can be estimated from marker pixels on discriminative regions of a few representative images. This work proposes flyweight networks, hundreds of times lighter than lightweight models, for SOD by combining a FLIM encoder with an \textit{adaptive decoder}, whose weights are estimated for each input image by a given heuristic function. Such FLIM networks are trained from three to four representative images only and without backpropagation, making the models suitable for applications under labeled data constraints as well. We study five adaptive decoders; two of them are introduced here. Differently from the previous ones that rely on one neuron per pixel with shared weights, the heuristic functions of the new adaptive decoders estimate the weights of each neuron per pixel. We 
compare FLIM models with adaptive decoders for two challenging SOD tasks with three lightweight networks from the state-of-the-art, two FLIM networks with decoders trained by backpropagation, and one FLIM network whose labeled markers define the decoder's weights. The experiments demonstrate the advantages of the proposed networks over the baselines, revealing the importance of further investigating such methods in new applications. 
\noindent 
\end{abstract}

\begin{IEEEkeywords}
Salient Object Detection, FLIM, Adaptive Decoders, Decoding
\end{IEEEkeywords}

\section{Introduction}\label{sec:intro}
 
Salient Object Detection (SOD) methods~\citep{wang2015deep, zhao2015saliency, borji2019salient} can locate objects that stand out in an image and assign higher values to their pixels in a saliency map. They then binarize the map, showing a segmentation mask with the detected objects. Many works rely on deep learning methods, while a recent tendency is lightweight models that require much less computational resources~\citep {liu2021samnet,lin2022lightweight,liang2024meanet}. Training such models requires extensive human effort to create segmentation masks (data annotation) by object delineation. Fine-tuning the models to new applications may also require considerable human effort in data annotation. 

In this work, we investigate a recent methodology for SOD~\citep{joao2023flyweight, visapp24}, which combines a convolutional encoder trained from user-drawn markers on discriminative regions of a few representative images~\citep{de2020feature} with an \textit{adaptive decoder} that creates saliency maps by estimating its parameters for each input image. The methodology allows the construction of \textit{flyweight} SOD models, hundreds of times more efficient than lightweight networks, trained with minimum human effort in data annotation and without backpropagation. The paper proposes and evaluates new adaptive decoders using a graph-based delineation algorithm to segment the objects. 

\begin{figure}[!htbp]
\makebox[\linewidth][c]{%
\begin{tabular}{*{4}{c@{\hskip 0.5pt}}c}
\includegraphics[width=0.245\columnwidth]{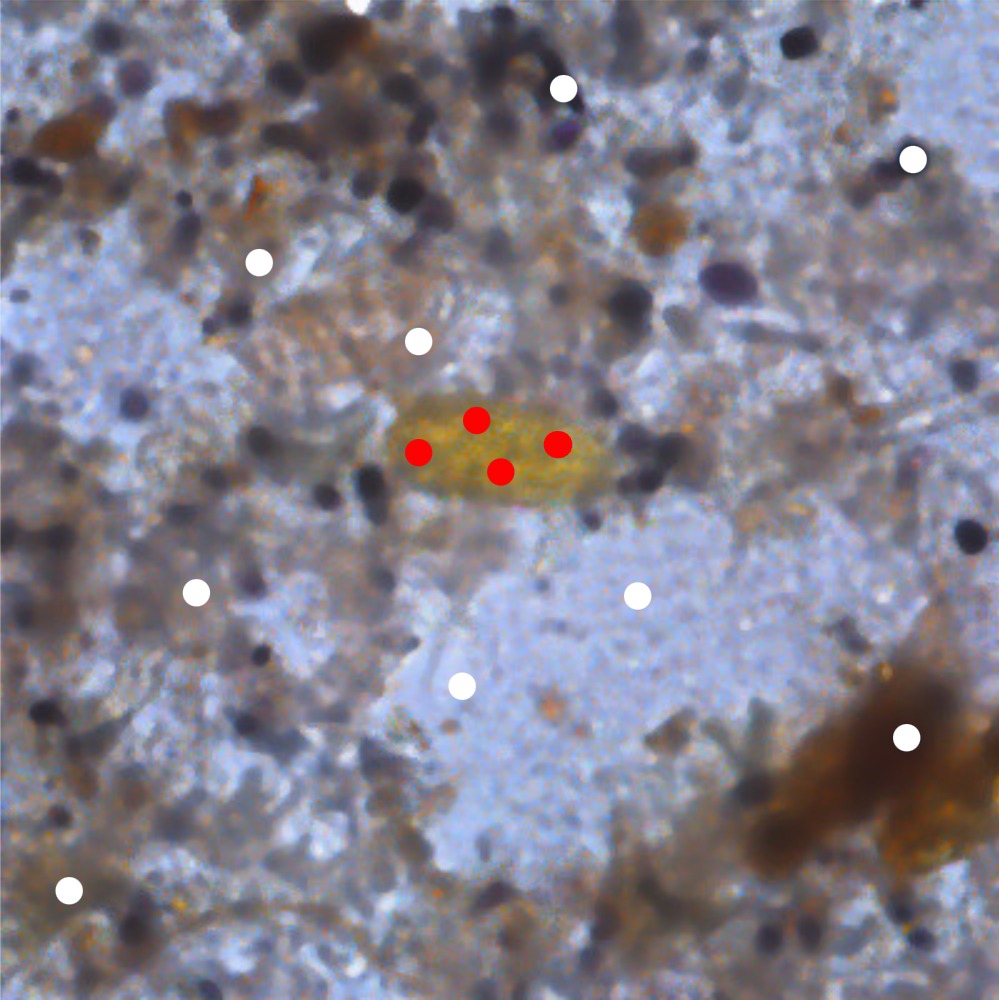} & 
\includegraphics[width=0.245\columnwidth]{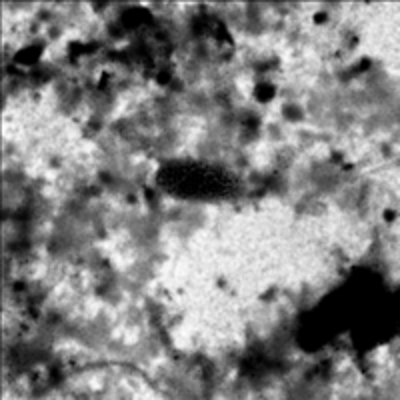} & 
\includegraphics[width=0.245\columnwidth]{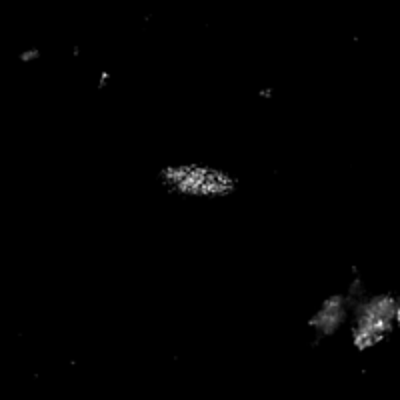} & 
\includegraphics[width=0.245\columnwidth]{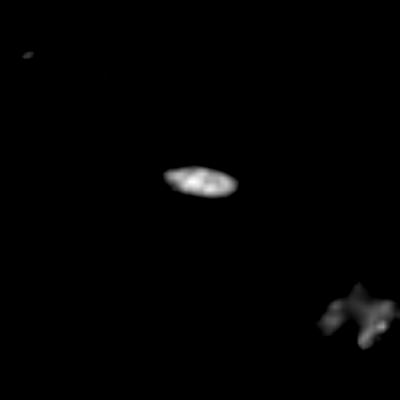} \\
(a) & (b) & (c) & (d)
\end{tabular}
}
\caption{Marker drawing for kernel estimation: (a) A training image with foreground (red) and background (white) markers; (b) background activation channel from a background kernel; (c) foreground activation channel from an object kernel; (d) resulting saliency map.}
\label{fig:feature_maps}
\end{figure}

Convolutional encoders are trained with \textit{Feature Learning from Image Markers} (FLIM). The user draws markers on discriminative (object and background) regions of representative images (\autoref{fig:feature_maps}a), and FLIM estimates the kernels of all convolutional blocks by clustering patches centered at marker pixels using the input feature map (activation channels) of each block and selecting the cluster's centers to constitute the kernels. A kernel derived from a background marker is expected to create a channel with background activation (\autoref{fig:feature_maps}b), while a kernel derived from an object marker is expected to generate a foreground activation channel (\autoref{fig:feature_maps}c). An adaptive decoder may be a simple point-wise convolution (i.e., a weighted average of the encoder's output channels) followed by activation, in which a heuristic function estimates the convolutional weights for each input image. The above characteristics of a FLIM encoder allow adaptive decoders to create reasonable saliency maps (\autoref{fig:feature_maps}d). 

\begin{figure}[!htbp]
\begin{center}
\begin{tabular}{cc}
\includegraphics[width=0.45\linewidth]{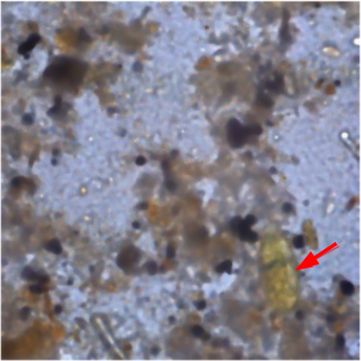} & 
\includegraphics[width=0.45\linewidth]{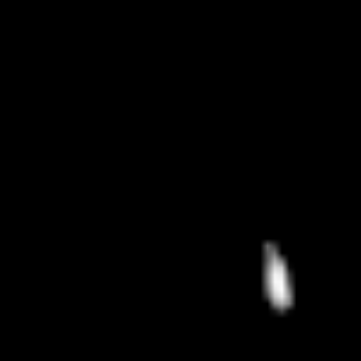}\\
(a) & (b)
\end{tabular}
\end{center}
\caption{A FLIM-based SOD model generalizing to new images. (a) A test image where the red arrow indicates the object (a parasite egg) and (b) the resulting saliency map.}
\label{fig:feature_maps_compl}
\end{figure}

In applications with redundant object and background properties, FLIM-based SOD models can be trained with only five images~\citep{joao2023flyweight} and generalize to new ones (Figures~\ref{fig:feature_maps_compl}a and~\ref{fig:feature_maps_compl}b). Moreover, FLIM networks have shown promising results in other tasks, such as object delineation~\citep{de2020feature}, image classification~\citep{de2020learning}, and instance segmentation~\citep{cerqueira2023building}, surpassing the same architecture trained from scratch and, in some situations, deep models~\citep{de2020learning,joao2023flyweight}.

This work studies five adaptive decoders: (i) a tri-state decoder, which is a slight modification of the original one in~\cite{visapp24}, (ii) an attention-based decoder, (iii) a label-based tri-state decoder, (iv) a probability-based decoder, and (v) a mean-based decoder. The present paper is an extended version of our previous work~\citep{soares2024adaptive}, in which decoders (i)-(iii) were introduced. Therefore, decoders (iv) and (v) are new contributions proposed here. The work compares FLIM-based SOD networks using the five adaptive decoders and three decoders with fixed weights (two of them learned by backpropagation). This comparative analysis also includes three lightweight SOD models from the state-of-the-art. The models were evaluated on two SOD tasks: (i) detection of parasite eggs in microscopy images, and (ii) brain tumor detection in magnetic resonance images. Given that recent lightweight networks adopt a delineation loss to improve saliency maps and segment the objects by thresholding, we explore a graph-based delineation algorithm for the same purpose. Graph-based object delineation was adopted only for parasite egg detection and was not used in~\cite{soares2024adaptive}.  

The experiments differ in several aspects from the ones in~\cite{soares2024adaptive}. FLIM-based SOD networks were created by two users, who followed an image selection procedure on a validation set to guarantee representative training images. The datasets were randomly divided three times into two parts with $50\%$ of the images each; one part was used to select the training set, leaving the remaining images for validation, and the other was used for test. Three to four representative training images were selected per split. The previous work used only a single test set per dataset and five training images chosen by a single user based on experience. As the work focuses on the decoder part, the encoder's architecture was selected using the validation set since an unsuitable encoder can negatively impact the decodification process. Another difference is the rule that defines the number of filters per block. Rather than specifying an empirical number of kernels per block, which required reducing the total number of kernels derived from the markers,  this work sets the number of kernels per block as the number of markers multiplied by the number of kernels per maker.

In summary, the contributions of this paper are: (1) two new adaptive decoders that introduce the concept of estimating one adaptive neuron per pixel for each image; (2) FLIM-based flyweight SOD networks that are hundreds of times more efficient than the current lightweight ones, and require very few images (three-four) to be trained, assuming the selected images are representative; (3) more extensive evaluation experiments involving two users, five adaptive decoders, six baselines, two datasets, and three metrics; and (4) the use of a graph-based algorithm for object delineation. The paper is organized in the following way: Related works are discussed in Section~\ref{sec:related_work}. Section~\ref{sec:theoretical_background} provides the background on FLIM encoders, while Section~\ref{sec:methods} presents the adaptive decoders. Section~\ref{sec:setup} describes the experimental setup, including the image selection and object delineation procedures.  Results are presented and discussed in Section~\ref{sec:results}. Finally, Section~\ref{sec:conclusion} states the conclusion and discusses future work. 

\section{Related Work}\label{sec:related_work}

Traditional SOD methods rely on local image features (texture, color, orientation) and a bottom-up approach to combine them into a saliency map~\citep{ullah2020brief}. One of the first works decomposes an image into feature maps that are further processed and combined at various scales to generate a saliency map ~\citep{itti2002model}. A similar strategy rescales the feature maps to the same domain of the input image before combining them into a saliency map with better resolution ~\citep{achanta2008salient}. There are also methods based on focusness~\citep{cheng2014global, zhu2014saliency} and objectness priors~\citep{chang2011fusing, jiang2013salient}.

Most recent SOD methods use a convolutional encoder for feature map extraction, due to its innate capacity to generate multi-level and multi-scale features~\citep{borji2019salient}. In the stage of feature map combination, methods fall into two categories: the ones that rely on multi-layer perceptrons (MLPs) for pixel classification into object/background~\citep{lee2016deep, he2015supercnn}, and those that generate a saliency map using a convolutional decoder~\citep{liu2018picanet, qin2019basnet,qin2020u2}. The second category, Fully Convolutional Networks -- FCNs, is the most common. PiCANet~\citep{liu2018picanet}, for instance, resembles the U-Net model with a convolutional encoder from the Resnet/VGG network. BASNet~\citep{qin2019basnet} adopts a similar U-shape with the encoder of ResNet-34 as the backbone and with a focus on boundary refinement through supervision in different layers of the network, including 
a module to refine the saliency map.
The idea was further enhanced by employing nested U-shaped structures across different network levels, enabling the capture of richer contextual information at varying scales~\citep{qin2020u2}. Unlike previous deep learning-based approaches, this method does not rely on a pretrained backbone and is instead trained from scratch.

Although the deep FCNs achieve state-of-the-art results, they consume considerable computational resources. In this scenario, lightweight models emerge while maintaining equivalent results. The methods can create a lightweight network by design or compression techniques -- e.g., knowledge distillation, quantization, and network simplification~\citep{chen2024review}. The former replaces convolution by group, separable, or dilated convolutions. These modified operations reduce the number of parameters and the computational complexity of the model. The expert usually does such changes. Still, it can be done automatically based on AutoML~\citep{chen2024review}, which searches the best architecture for the network~\citep{he2021automl} -- a subtopic also known as Network Architecture Search (NAS).

 \begin{sloppypar}
Lightweight models are usually pre-trained on ImageNet~\citep{deng2009imagenet}. Several models~\citep{lin2022lightweight,li2023lightweight,liang2024meanet} use the encoder of MobileNet-V2~\citep{sandler2018mobilenetv2} as the backbone to constrain the model's size. Two well-succeeded examples are MSCNet~\citep{lin2022lightweight} and MEANet~\citep{liang2024meanet}, both developed to cope with the multi-scale information of salient objects in remote sensing images. To address the problem, MSCNet and MEANet introduce a multi-scale context extraction module and a multi-scale edge-embedded attention module. Other works developed custom backbones~\citep{wang2023larnet, wang2023elwnet, zhou2024admnet, liu2021samnet}. They argue that utilizing pre-trained lightweight encoders from MobileNet-V2~\citep{sandler2018mobilenetv2}, ShuffleNet~\citep{zhang2018shufflenet}, and GhostNet~\citep{han2020ghostnet} makes it more difficult to compress the model, generates redundant features, and affects the quality of the saliency maps. One well-succeeded example is SAMNet ~\citep{liu2021samnet}, which introduces a stereoscopically attentive multi-scale module to fuse image features at various scales. 
\end{sloppypar}

Among lightweight networks that do not rely on pre-trained backbones, FLIM-based models~\citep{de2020learning} have proven effectiveness in detecting~\citep{joao2023flyweight}, classifying~\citep{de2020learning}, and segmenting objects~\citep{de2020feature,cerqueira2023building}. In all methods, the encoder is trained from user-drawn markers without backpropagation. However, backpropagation may improve the encoder when annotated examples are available in a reasonable number~\citep{cerqueira2023building}. For SOD tasks, FLIM-based networks use adaptive decoders ~\citep{joao2023flyweight, visapp24, soares2024adaptive}, which allow training the entire model from a few marked images without backpropagation. 

This work introduces two adaptive decoders besides the three decoders presented in~\cite{soares2024adaptive} and adds a graph-based delineation algorithm to segment the object. Alternatively, one could improve the saliency map before segmentation by exploring cellular automata~\citep{salvagnini2024improving}. We define the number $m'=m_1\times m_2$ of kernels per block as the number $m_1$ of markers multiplied by the number $m_2$ of kernels per marker. Previous works define $m' < m_1\times m_2$ requiring methods that reduce to $m'$ the total number $m_1\times m_2$ of estimated kernels, which may miss important information.

An essential step in FLIM-based networks is selecting representative training images. We wish to choose only a few representative examples from an unlabeled set to minimize human effort in data annotation. Supervised ~\citep{cerqueira2024interactive2} and unsupervised ~\citep{cerqueira2024interactive} approaches are under investigation, and such representative images are crucial for the encoder's performance. Since the decoder is our focus, we adopt a supervised approach based on \cite{cerqueira2024interactive}, with the difference that we make the selection based on the result after the decoder, not on the feature maps and do not manually annotate the convolutional filters. We select one image at a time and update the model to verify an incremental performance on a validation set. The user retains the images that improve the model and stops with three or four images in the training set.

\section{FLIM encoders}\label{sec:theoretical_background}

This section provides definitions related to FLIM encoders. The basic idea is to estimate the convolution kernels from image patches, by using $k$-means clustering on a set of patches located around some markers selected by the user. To remove bias, the set of patches undergo a marker-based normalization process.

\subsection{Images and patches}

Let $\textbf{I}$ be an image with~$m$ channels and domain~$D_I\subset \mathcal{Z}^2$ of size~$w\times h$ pixels, such that~$\vec{I}(p) = (I_1(p),I_2(p),\ldots,I_m(p))\in \mathbb{R}^m$ assigns~$m$ features to every pixel~$p = (x_p,y_p)\in D_I$. Let~$A(p)$ be a set of~$k \times k$ adjacent pixels~$q=(x_q,y_q)\in D_I$, $x_q-x_p\in [-d\frac{k}{2},d\frac{k}{2}]$ and~$y_q-y_p\in [-d\frac{k}{2},d\frac{k}{2}]$, within a squared region of size~$d k \times d k$ centered at~$p$ with dilation factor~$d\geq 1$. A patch~$\vec{P}(p)\in \mathbb{R}^{k\times k\times m}$ results from the concatenation of feature vectors~$\vec{I}(q)$ of all~$q\in A(p)$. 

\subsection{Sets of marker pixels}

Let $\mathcal{T}$ be a set of training (representative) images with $m$ channels, $\mathcal{M}(\textbf{T})$ be the set of all marker pixels drawn on image $\textbf{T}\in \mathcal{T}$, and $\mathcal{M} = \bigcup_{\textbf{T}\in \mathcal{T}} \mathcal{M}(\textbf{T})$.

\subsection{Convolutional block and FLIM encoder}

A convolutional block may contain different image operations. We consider marker-based normalization, convolution, activation, and pooling. A FLIM encoder is a sequence of convolutional blocks $b=1,2,\ldots,B$. 

\subsubsection{Marker-based normalization}

Let $\textbf{I}^{0}$ with $m$ channels be an input image of the first block $b=1$. Each channel $I^{0}_j$, $j=1,2,\ldots,m$, of
$\textbf{I}^{0}$ can be normalized by

\begin{align}
I^{0}_j(p) &\leftarrow \frac{I^{0}_j(p)-\mu_j}{\sigma_j + \epsilon},\\
\mu_j &= \frac{1}{|\mathcal{M}|}\sum\limits_{\textbf{T}\in\mathcal{T},p\in\mathcal{M}(\textbf{T})} T_j(p),\\
\sigma^2_j &= \frac{1}{|\mathcal{M}|}\sum\limits_{\textbf{T}\in\mathcal{T},p\in\mathcal{M(\textbf{T})}}(T_j(p)-\mu_j)^2,
\end{align} 
\noindent
where $\epsilon > 0$ is a small value. This operation centralizes the patches around the origin of $\mathbb{R}^{k\times k\times m}$ and corrects distortions among different features, dismissing the need for estimating bias. When $\textbf{I}^{b-1}$ is the input image of a deeper block $b > 1$, the sets $\mathcal{M(\textbf{T})}$ of marker pixels must be mapped to the domain of image $\textbf{I}^{b-1}$.  

\subsubsection{Convolution, activation, and pooling}

Convolution kernels are estimated from image patches. Let $\mathcal{P}(\textbf{T})$ be the set of patches from the marker pixels of a given marker on a training image $\textbf{T}\in \mathcal{T}$. The patches in $\mathcal{P}(\textbf{T})$ are obtained after marker-based normalization. For a given number of kernels per marker, we run the $k$-means clustering on the patch dataset $\mathcal{P}(\textbf{T})$ to obtain one kernel $\vec{K} \in \mathbb{R}^{k\times k\times m}$ with norm $\|\vec{K}\|=1$ from each cluster center. 

In this work, the number $m'$ of kernels in a given block is the total number of kernels per marker multiplied by the number of markers, considering all training images in $\mathcal{T}$. The above kernel estimation procedure then obtains $m'$ kernels $\vec{K}_j$, $j=1,2,\ldots,m'$. Although marker labels may be explored for kernel estimation~\citep{de2020learning}, this option is not adopted in this work. 

The convolution between the input image $\textbf{I}^{b-1}$ and each kernel $\vec{K}_j$ generates an image $\textbf{J}$ with $m'$ channels $J_j$, such that
\begin{eqnarray}
J_j(p) & = & \langle \vec{P}(p), \vec{K}_j \rangle,
\end{eqnarray}
$j=1,2,\ldots,m'$. We apply ReLU followed by max-pooling to each channel, such that the output image $\textbf{I}^{b}$ of the block has $m'$ activation channels.

By drawing markers with labels $l\in \{1,2\}$ (background, foreground) and estimating kernels from each marker, the label $\lambda(I^{b}_j)\in \{1,2\}$
of channel $I^{b}_j$ is the same of the marker used to estimate the kernel $\vec{K}_j$ that has generated it. This is valid for any block $b\in \{1,2,\ldots,B\}$ and one can expect background and foreground activations in the respective channels. However, this might not occur due to the cross-influence among training images. For this reason, adaptive decoders rely on a heuristic function that estimates background and foreground activation channels for each input image.

\section{Adaptive Decoders}\label{sec:methods}

\begin{figure}[!htbp]
\begin{center}
\begin{tabular}{cc}
\includegraphics[width=0.48\columnwidth]{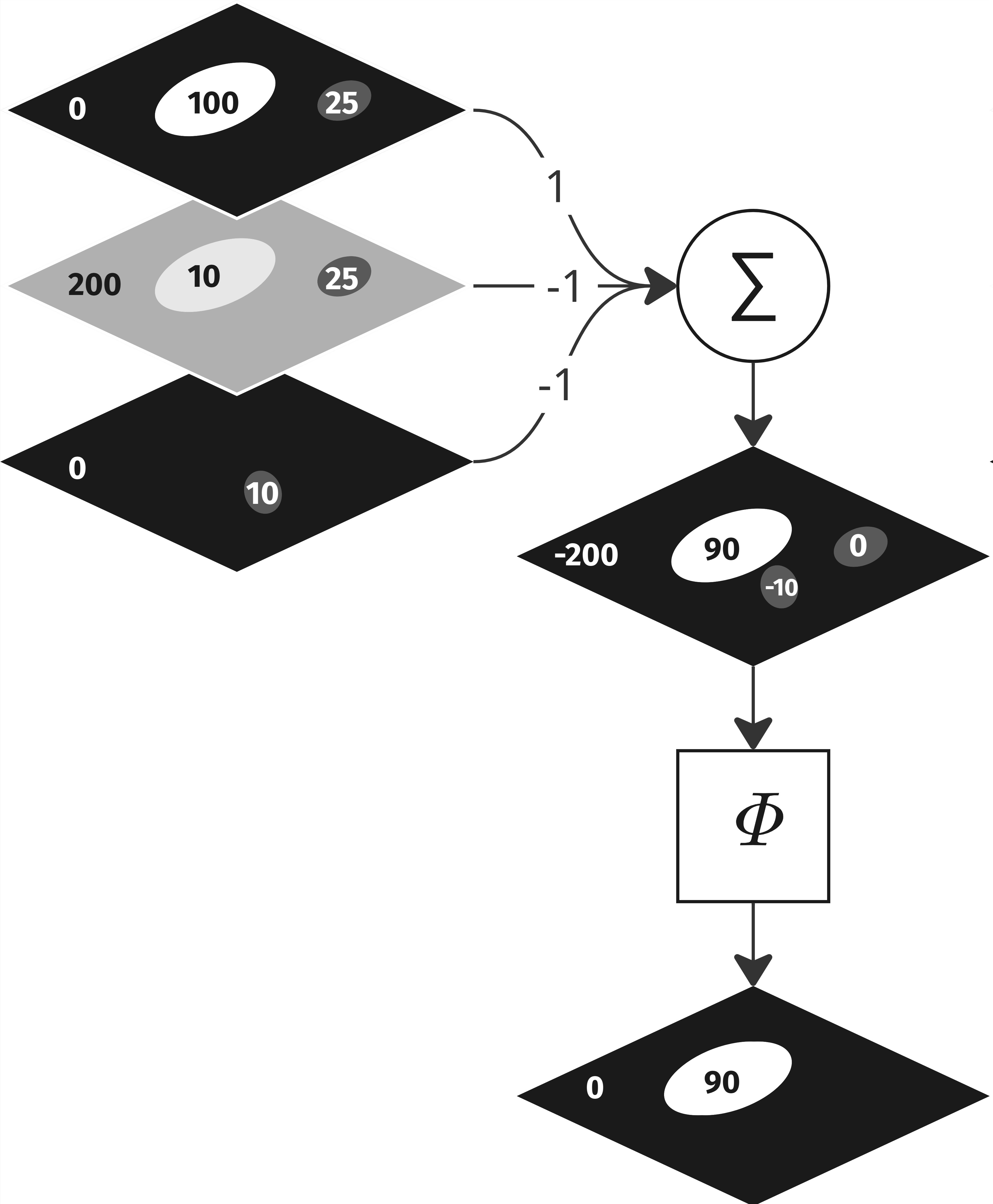} & 
\includegraphics[width=0.48\columnwidth]{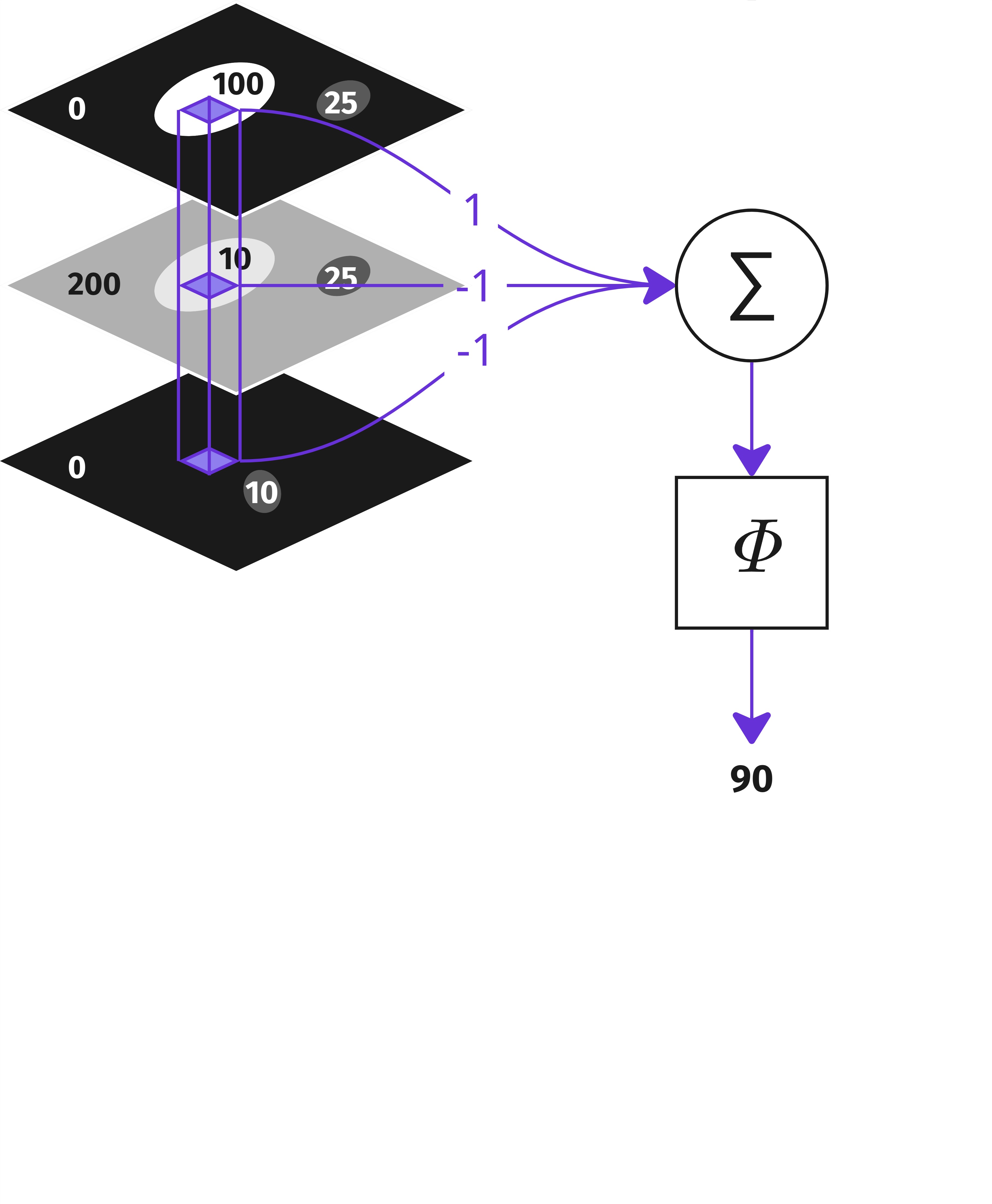}\\
(a) & (b)
\end{tabular}
\end{center}
\caption{Adaptive decoders. (a) A point-wise convolution followed by activation, whose weights are estimated in ${-1,+1}$ by a heuristic function. (b) It defines one neuron per pixel, and all neurons share the same weights.}
\label{fig:adaptive_decoder}
\end{figure}

\begin{sloppypar}
An adaptive decoder is a method that can combine the activation channels (image features) from the feature map of a given block into an object saliency map by adapting its parameters for each input image. The simplest example is a point-wise convolution between the feature map $\textbf{I}^{L}$ and a point-wise kernel $\vec{\alpha}=(\alpha_1,\alpha_2,\ldots,\alpha_{m'})\in \mathbb{R}^{1\times 1\times m'}$, followed by ReLU activation $\Phi$ (Figure~\ref{fig:adaptive_decoder}a-b). Such a decoder represents one neuron per pixel, with shared weights estimated by a given heuristic function. One may also re-estimate the weights of the point-wise convolution for each pixel.

In SOD, FLIM encoders are designed to either activate or deactivate object pixels in different channels. Although this expected behavior is not guaranteed, the presented adaptive decoders can create a reasonable saliency map $\textbf{S}$ with image domain $D_S\subseteq D_I$ and values $S(p)\in \mathbb{R}$ for each $p\in D_S$, such that
\begin{eqnarray}
S(p) & = & \Phi\left(\langle \vec{I}^L(p), \vec{\alpha}\rangle\right)=\Phi\left(\sum_{i=1}^{m'} \alpha_i  I_i^L(p)\right).
\label{eq.decoder}
\end{eqnarray} 
One may upsample $D_S=D_{I^L}$ to the size of $D_{I^0}=D_I$, whenever strides greater than one are used. 
\end{sloppypar}

\begin{figure}[htb!]
\begin{center}
\newcommand\sizefig{0.33\columnwidth}
\centering
\makebox[\columnwidth][c]{%
\renewcommand{\arraystretch}{1}
\begin{tabular}{*{3}{c@{\hskip 0.5pt}}c}
\includegraphics[width=\sizefig]{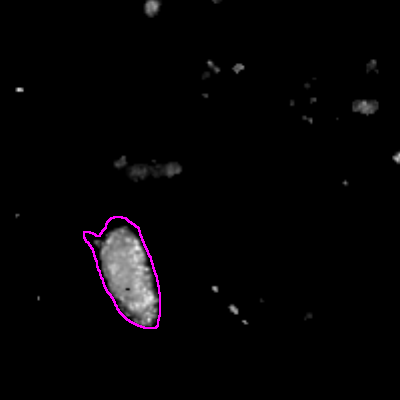}&
\includegraphics[width=\sizefig]{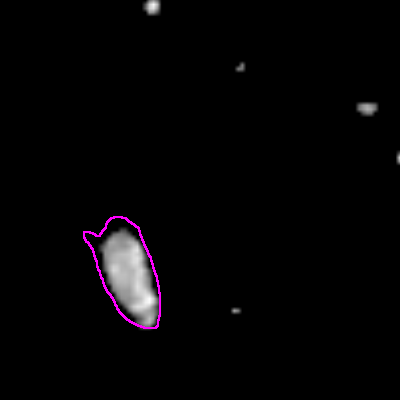}&
\includegraphics[width=\sizefig]{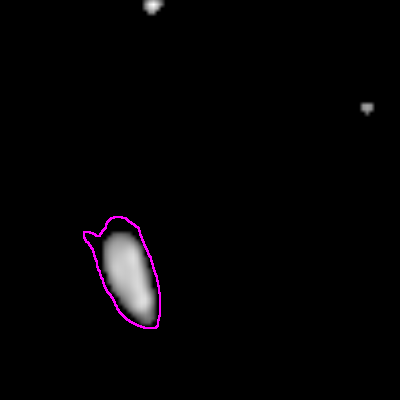}\\
\footnotesize(a) Block 1 & \footnotesize(b) Block 2 & \footnotesize(c) Block 3\\
\end{tabular}
}
\caption{Saliencies generated by an adaptive decoder for different blocks of an architecture. The ground-truth's border is presented in magenta.}
\label{fig:delineation_result_blocks}
\end{center}
\end{figure}

An interesting effect of combining a FLIM encoder with an adaptive decoder is that the saliency maps after each encoder's block show fewer false positives as the encoder adds blocks. However, object representation is more precise in the earlier blocks (Figure~\ref{fig:delineation_result_blocks}). 

Next, we describe the proposed adaptive decoders. Differently from the previous ones in ~\cite{soares2024adaptive}, the new adaptive decoders, probability-based and mean-based, adopt heuristic functions that estimate weights for each neuron per pixel. Hence, the neurons do not share weights. 

\subsection{Tri-state adaptive decoder (\texorpdfstring{$ts$}{})}

A tri-state adaptive decoder assumes the object occupies a considerably smaller portion of the image domain compared to the background. It identifies an object (background) activation channel when its mean activation is below (above) a threshold $\tau - \sigma$ ($\tau + \sigma$), and when the fraction of pixels above Otsu threshold at the channel is below (above) another threshold $t_i$. Such rules define each channel weight $\alpha^{ts}_i$, with $\alpha^{ts}_i=0$ when the rules are not satisfied. 

Let $\mu_{I_i^L}$ be the mean activation of channel $I_i^l$, with $\tau$ denoting the Otsu threshold for the distribution $\{\mu_{I_1^L},\mu_{I_2^L},\ldots,\mu_{I_{m'}^L}\}$, and $\sigma$ indicating the standard deviation of that distribution. The number of pixels exceeding the Otsu threshold for channel $I_i^L$ divided by $|D_{I}|$ establishes a threshold $t_i$. The \textit{tri-state adaptive decoder} ($ts$) defines $\alpha^{ts}_i$ as follows.

\begin{equation}\label{eq:w_adaptive}
\alpha^{ts}_i = \begin{cases}
    -1,      & \text{if}  \quad \mu_{I_i^L} \geq \tau + \sigma \mbox{ and } t_i > 0.2,\\
    +1,       & \text{if} \quad \mu_{I_i^L} \leq \tau - \sigma \mbox{ and } t_i < 0.1,\\
    0,       & \text{otherwise}.
\end{cases}
\end{equation}

The limits for the threshold, $t_i > 0.2$ and $t_i < 0.1$, are empirically defined for the datasets utilized in this work.

\subsection{Attention-based adaptive decoder (\texorpdfstring{$at$}{})}

In \cite{li2022cam}, the authors introduce a method to calculate channel attention without utilizing trainable parameters. Channel attention should assign positive weights (more importance) to object channels and negative weights (less importance) to background channels. In our work, we adopt a slight variation of their technique. Let $X$ and $Y$ represent the outcomes of max-pooling and average pooling through all channels $I_i^L$, $i\in[1, m']$. $X$ and $Y$ have their values linearly scaled to  $[0,1]$. A spatial attention $a$ is established by linearly normalizing the sum of $X$ and $Y$ within $[0,1]$. Denote $\vec{a} \in \mathbb{R}^{w\times h}$ as the vectorized form of the spatial attention $a$, and $\vec{b}_i\in \mathbb{R}^{w\times h}$ as the vectorized form of channel $I_i^L$. Each channel's importance $c_i$ is calculated by 
\begin{eqnarray} 
 c_i & = & \frac{\langle \vec{a}, \vec{b}_i\rangle}{\|\vec{a}\|\|\vec{b}_i\|}.
\end{eqnarray}

Let $\mu_c$ and $\sigma_c$ be the mean and standard deviation of the distribution $\{c_1, c_2,\ldots, c_{m'}\}$. The weight vector for an attention-based decoder can be expressed as $\vec{\alpha}^{at}=(\alpha^{at}_1,\alpha^{at}_2,\ldots,\alpha^{at}_{m'})$, where

\begin{eqnarray}
\alpha^{at}_i & = & \begin{cases}
+1, & \mbox{if $c_i < \mu_c - \frac{\sigma_c}{2}$,}\\
-1, & \mbox{if $c_i > \mu_c + \frac{\sigma_c}{2}$}\\
0, & \mbox{otherwise}
\end{cases}
\end{eqnarray}
for $i\in [1,m']$.

\subsection{Label-based tri-state adaptive decoder (\texorpdfstring{$lt$}{})}
Assuming that object and background activation channels are derived from kernels according to their labeled markers, the label-based tri-state adaptive decoder combines evidences from this information and the tri-state adaptive decoder. Let $\vec{\alpha}^{ts}=(\alpha^{ts}_1,\alpha^{ts}_2,\ldots,\alpha^{ts}_{m'})$ be the weight vector generated by the tri-state decoder for a particular input image, and $\lambda(I^L_i)\in \{1,2\}$, $i\in [1,m']$, be the label for object and background activation channels. We establish
\begin{eqnarray}
\alpha^{lt}_i & = & \begin{cases}
0, & \mbox{if $\lambda(I^L_i)=2$,}\\
\alpha^{ts}_i, & \mbox{otherwise}.
\end{cases}
\end{eqnarray}
Hence, we only consider the weights of the tri-state adaptive decoder for foreground-labeled kernels, eliminating channels where background-labeled kernels generate object activations.

\subsection{Probability-based adaptive decoder (\texorpdfstring{$pb$}{})}

Let $\mathcal{A}(p)$ be a small neighborhood around $p$, $\mu_1(p)$ be the mean activation value of the adjacent pixels $q \in \mathcal{A}(p)$ in all the channels with $\lambda(I^L_i) = 1$, and $\mu_2(p)$ be the mean activation of the adjacent pixels $q\in\mathcal{A}(p)$ in all channels with $\lambda(I^L_i)=2$. The values $\mu_j(p), j={1,2}$ are:

\begin{equation}
    \mu_j(p) = \frac{1}{N_j} \sum_{i,\lambda(I^L_i)=j}\sum_{q\in\mathcal{A}(p)}I^L_i(q),
\end{equation}

\noindent where $N_j$ is the total number of adjacent pixels of $p$ in channels with label $j$. Similarly, the variance $\sigma^2_j(p), j={1,2}$ is:

\begin{equation}
    \sigma_j^2(p) = \frac{1}{N_j} \sum_{i,\lambda(I^L_i)=j}\sum_{q\in\mathcal{A}(p)}(I^L_i(q) - \mu_j(p))^2.
\end{equation}

We estimate the probability of $p$ be part of the object (background) given its activation $I^L_i(p)$ in a given channel by:
\begin{equation}
    \phi_{i,j}(p) = \exp^{\left(\frac{(I^L_i(p) - \mu_j(p))^2}{2\sigma^2_j(p)}\right)}.
\end{equation}

The probability-based adaptive decoder assigns  a weight vector $\vec{\alpha}^{pb}(p)=(\alpha^{pb}_1(p),\alpha^{pb}_2(p),\ldots,\alpha^{pb}_{m'}(p))$ per pixel rather than a single weight vector per channel, as the previous ones do. For each channel, it assumes consistency between the channel's label and those probability values. 
\begin{align}
\alpha^{pb}_i(p) & = & \begin{cases}
+1, & \mbox{if $\lambda(I_i^L(p)) = 1$ and $\phi_{i,1}(p) > \phi_{i,2}(p)$},\\
-1, & \mbox{if $\lambda(I_i^L(p)) = 2$ and $\phi_{i,1}(p) < \phi_{i,2}(p)$},\\
0, & \mbox{otherwise}.
\end{cases}
\end{align}

\subsection{Mean-based adaptive decoder (\texorpdfstring{$mb$}{})}

The mean-based decoder is a simplification of the probability-based adaptive decoder for the sake of efficiency, assuming that 
$\mu_1(p) > \mu_2(p)$ for object pixels and $\mu_1(p) < \mu_2(p)$ for background pixels, being this information consistent with the channel's label. The weight vector $\vec{\alpha}^{mb}(p) = (\alpha_1^{mb}(p),\alpha_1^{mb}(p),\cdots, \alpha_{m'}^{mb}(p))$ is defined by:

\begin{align}
\alpha^{mb}_i(p) & = & \begin{cases}
+1, & \mbox{if $\lambda(I_i^L(p)) = 1$ and $\mu_1(p) > \mu_2(p)$},\\
-1, & \mbox{if $\lambda(I_i^L(p)) = 2$ and $\mu_1(p) < \mu_2(p)$},\\
0, & \mbox{otherwise.}
\end{cases}
\end{align}

\section{Experimental setup}\label{sec:setup}

\begin{figure*}[htb!]
\centering
\includegraphics[width=\textwidth]{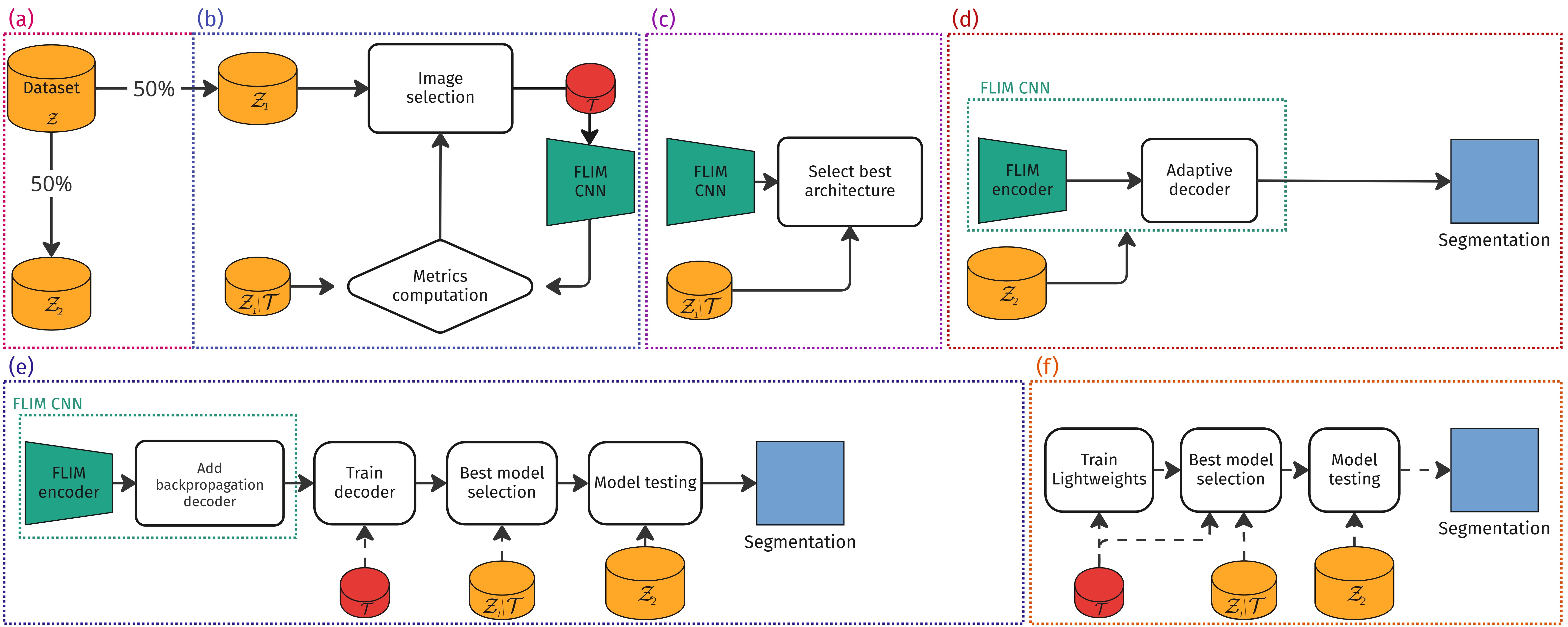}
    \caption{Pipeline used for the experiments. (a) The dataset $\mathcal{Z}$ is randomly divided into sets $\mathcal{Z}_1$ and $\mathcal{Z}_2$. (b) A few representative images are selected for the set $\mathcal{T}$, according to the model's performance on the validation set $\mathcal{Z}_1\backslash \mathcal{T}$. (c) Once $\mathcal{T}$ is fixed, the best network architecture is found on the validation set. (d) The pre-trained FLIM encoder with each adaptive decoder is tested on $\mathcal{Z}_2$. (e) Each fixed-weight decoder is trained on $\mathcal{T}$, using the pre-trained FLIM encoder as the fixed backbone; the best models are found on $\mathcal{Z}_1\backslash \mathcal{T}$ and tested on $\mathcal{Z}_2$. (f) Each pre-trained lightweight model is fine-tuned on $\mathcal{T}$; the best models are found on $\mathcal{Z}_1\backslash \mathcal{T}$ and tested on $\mathcal{Z}_2$. }
\label{fig:pipeline}
\end{figure*}

\autoref{fig:pipeline} illustrates the experimental setup.  First, a dataset $\mathcal{Z}$ is randomly divided into two parts, $\mathcal{Z}_1$ and $\mathcal{Z}_2$, each with 50\% of the samples (\autoref{fig:pipeline}a). $\mathcal{Z}_1$ is used to select the training set $\mathcal{T}$ image by image as the model (a FLIM CNN) is trained on $\mathcal{T}$, evaluated on $\mathcal{Z}_1\backslash \mathcal{T}$, and a new image, among the ones not solved by the model, is selected for $\mathcal{T}$ (\autoref{fig:pipeline}b). Once $\mathcal{T}$ is defined with 3 or 4 images, depending on the split and user,  the best CNN architecture on $\mathcal{Z}_1\backslash \mathcal{T}$ is selected (\autoref{fig:pipeline}c). Its FLIM encoder and different adaptive decoders form the FLIM CNNs for testing on $\mathcal{Z}_2$ (\autoref{fig:pipeline}d). 
The set $\mathcal{T}$ is also used to train the decoders of two FLIM CNNs (\autoref{fig:pipeline}e) and fine-tune the lightweight models by backpropagation (\autoref{fig:pipeline}f), subsequently using the set $\mathcal{Z}_1\backslash \mathcal{T}$ to select the best model for testing on $\mathcal{Z}_2$. At every stage, the saliency maps are post-processed to produce predicted binary masks and evaluate their metrics. The whole process is repeated three times for each user and dataset, with random splits of $\mathcal{Z}$, to obtain the mean and standard deviation of the metrics.  

The following sections describe the datasets, the best architectures of the FLIM CNNs with adaptive decoders, the baselines, the evaluation metrics, the representative image selection procedure, and the post-processing of the saliency maps to obtain the predicted binary masks.

\subsection{Datasets}

Two SOD tasks were used to evaluate the methods: parasite egg detection from optical microscopy images and brain tumor detection from magnetic resonance (MR) images. The first task uses a dataset, Parasites, of \textit{Schistosoma Mansoni} eggs (see availability of data and materials). The second task uses 2D grayscale slices from the Brain Tumor Segmentation Challenge 2021, BraTS, a public dataset~\citep{menze2014multimodal, bakas2017advancing, bakas2018identifying}. A challenge in the Parasites is the high amount of food debris (impurities) in the images, some of which are without eggs, and others contain impurities similar to the eggs in shape and color. In the BraTS dataset, the challenge comes from tumors with higher intensity, size, and shape variance.  

Parasites contains $1,219$ RGB images with dimensions of $400\times400$ pixels. BraTS contains $3,743$ $16$-bit grayscale images with dimensions of $240\times240$ pixels.

\subsection{FLIM CNNs with adaptive decoders}

Each user specifies the hyperparameters of each encoder block $b=1,2,\ldots, B$ as consisting of patch sizes, the number of kernels per marker, kernel-dilation factors, the number of desired kernels, pooling adjacency, pooling strides, and pooling type. The kernels are estimated as described in Section~\ref{sec:theoretical_background}. The output of the last block is a feature map $\textbf{I}^{B}$ to the input of one of the adaptive decoders, forming the following CNNs: FLIM$_{ts}$, FLIM$_{at}$, FLIM$_{lt}$, FLIM$_{pb}$, and FLIM$_{mb}$. 

\begin{figure}[h!]
\begin{center}
\includegraphics[width=\columnwidth]{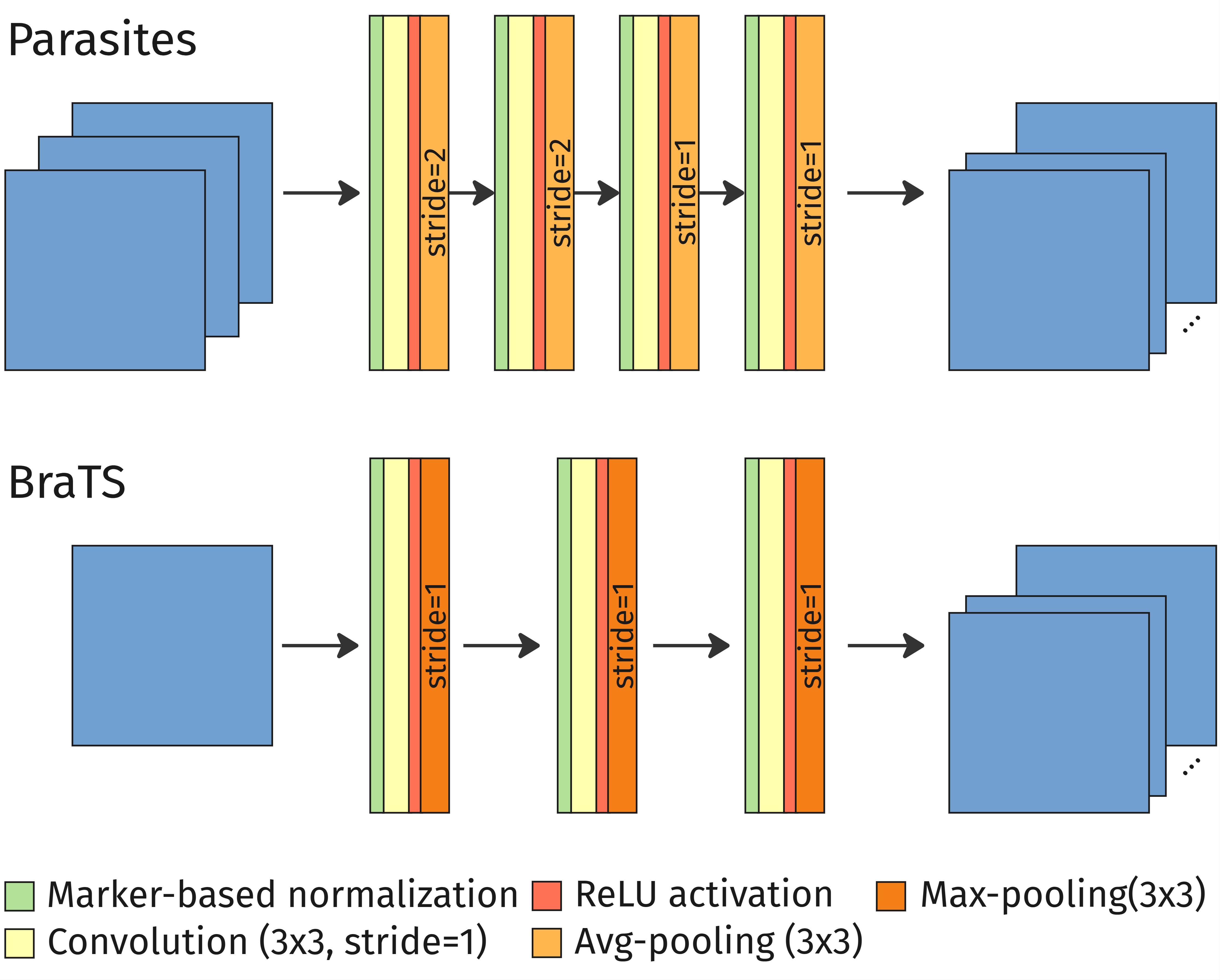}
\caption{Maximum-depth encoders evaluated for parasite egg and brain tumor detection, with the operators used in each block.}
\label{fig:flim_architecture}
\end{center}
\end{figure}

The number of kernels per block is the total number of markers multiplied by the number of kernels per marker. Hence, both users start from an encoder with four blocks ($B=4$) for parasite egg detection and three ($B=3$) for brain tumor detection, as shown in Figure~\ref{fig:flim_architecture}. Each block is composed of a marker-based normalization layer, convolution with kernel size of $3\times 3$, ReLU activation and pooling (average or max). Since the number of kernels per block varies for each user and split, the best CNN architecture per split is obtained by verifying the adaptive decoder and block $1\leq b \leq B$ that produces the best metric $F_\beta$ on $\mathcal{Z}_1\backslash \mathcal{T}$ (Section~\ref{sec:best_flim_arch}).  

\subsection{Baselines}

As baselines, we selected CNNs using the same FLIM encoder and substituting the adaptive decoders by decoders with fixed weights.  We also selected three lightweight models from the state-of-the-art as baselines.

\subsubsection{FLIM SOD models with fixed-weight decoders}

Two decoders with fixed weights are point-wise convolutions followed by ReLU activation, one of them trained by backpropagation. The third decoder forms a U-shape network with the FLIM encoder, named \texorpdfstring{U-Net$_{\text{FLIM}}$}{}, trained by backpropagation.

 \paragraph{Label-based fixed-weight decoder ($lm$)} This decoder assumes that a kernel from a labeled marker will create a channel with the same label. Hence, we can assume that $\lambda(I_i^L)\in \{-1, +1\}$ indicates background ($-1$) or foreground ($+1$) channel and define the weight vector of a label-based fixed-weight decoder as $\vec{\alpha}^{lm}=(\alpha^{lm}_1,\alpha^{lm}_2,\ldots,\alpha^{lm}_{m'})$, where $\alpha^{lm}_i=\lambda(I_i^L)$, $i=1,2,\ldots,m'$. The CNN with $lm$ is called FLIM$_{lm}$.

\paragraph{Backpropagation-based fixed-weight decoder ($bp$)} We fixed the pre-trained FLIM encoder and optimized the weight vector $\vec{\alpha}^{bp}=(\alpha^{bp}_1,\alpha^{bp}_2,\ldots,\alpha^{bp}_{m'})$ of the fixed-weight decoder by backpropagation using the ground-truth and predicted masks of the images in $\mathcal{T}$. The weight vector $\vec{\alpha^{bp}}$ is initialized with the Xavier method. The loss function is the average between  Dice and Binary Cross Entropy (BCE). Adam performed the optimization with a learning rate of 0.01 for 100 epochs. The CNN with $bp$ is called FLIM$_{bp}$.

\paragraph{\texorpdfstring{U-Net$_{\text{FLIM}}$}{}}

\begin{figure}[h!]
\begin{center}
\includegraphics[width=\columnwidth]{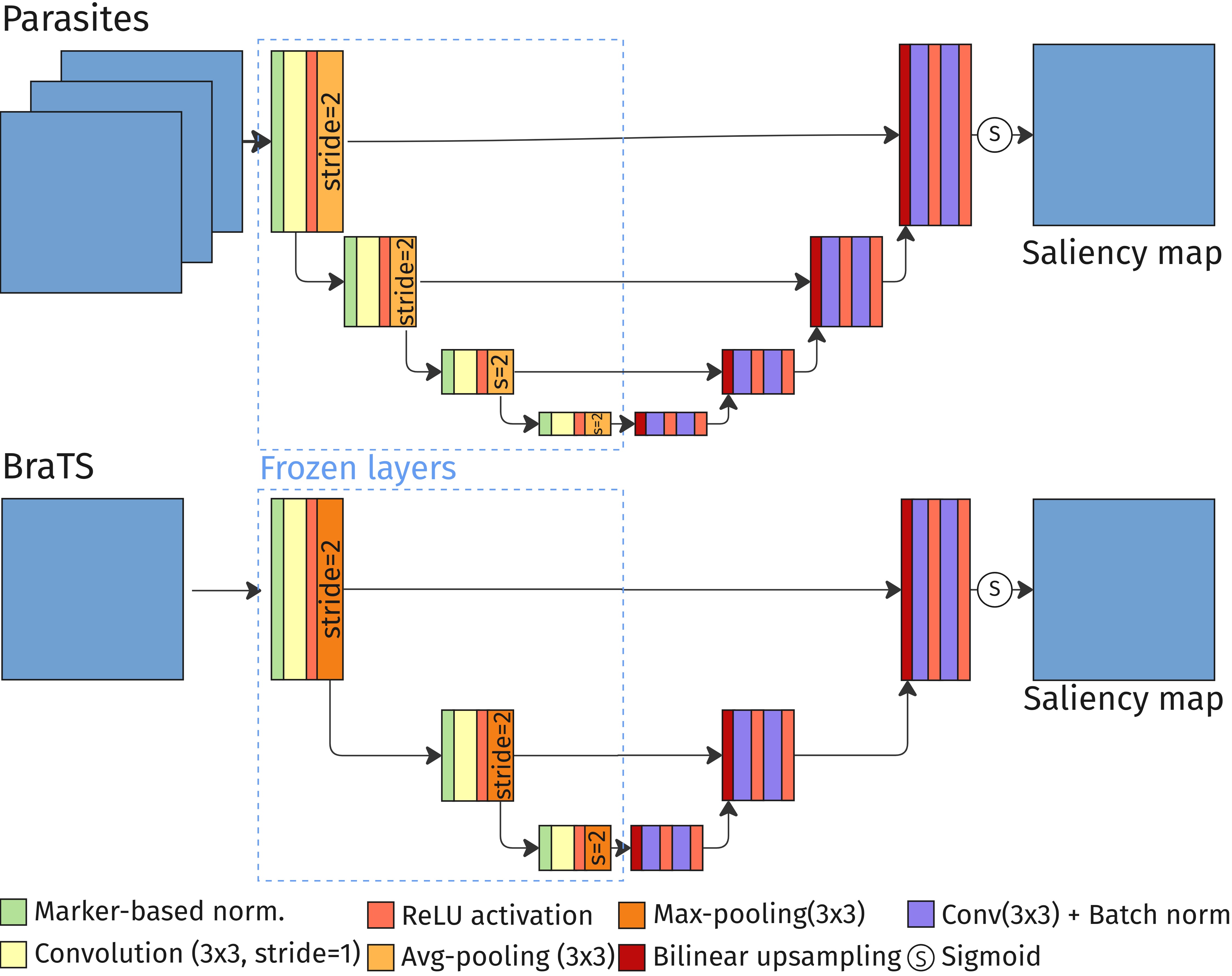}
\caption{\texorpdfstring{U-Net$_{\text{FLIM}}$}{} architectures with the operators used in each block.}
\label{fig:unet_architecture}
\end{center}
\end{figure}

This CNN uses the pre-trained FLIM encoder frozen, considers a U-shape with skip connections from the output of each encoder's block to each decoder's block, and trains the model on $\mathcal{T}$. The FLIM encoder was set with four blocks for the Parasites and three blocks for the brain tumor dataset (Figure~\ref{fig:unet_architecture}). They use the same architecture and weights from the FLIM base architecture from \autoref{tab:architectures}, with each block being composed of marker-based normalization, convolution with kernel size $3\times3$ and stride $1$, ReLU activation and pooling with size $3\times3$ and stride changed to $2$. The decoder was composed of blocks consisting of a bilinear upsampling, convolution with kernel size $3\times3$ and stride $1$, batch normalization, and ReLU activation. After the last block of the decoder, a sigmoid function was applied to generate the saliency map.

The decoder's blocks used bilinear interpolation for upsampling, convolution, and batch normalization. The models were trained for 50 epochs, with a learning rate linearly decreasing from 0.01 to 0.00001. Horizontal and vertical flips were used for data augmentation.

\subsubsection{Lightweight SOD models}

We used the codes provided by the authors for the lightweight SOD models,  SAMNet, MSCNet, and MEANet, with the default of almost all configurations. We used their pre-trained weights and fine-tuned them on set $\mathcal{T}$. The three models were trained for 100 epochs. For convergence, we had to increase their original learning rate. SAMNet, MSCNet, and MEANet were trained with learning rates of 0.005,  0.03, and  0.01, respectively. The batch size was set to $|\mathcal{T}|$. The best model for SAMNet was selected on $\mathcal{Z}_1\backslash \mathcal{T}$, but MSCNet and MEANet can only select their best model on $\mathcal{T}$.

\subsection{Evaluation metrics}\label{sec:metrics}

The resulting saliency maps of the models were post-processed (Section~\ref{sec:postproc}) to obtain predicted binary masks. We used Mean Absolute Error (MAE) and F-measure ($F_\beta$) to compare those masks with the ground-truth binary masks of the images. The parameter $\beta$ was set to $0.3$,  giving more weight to Precision, as it is usually done in the literature. Those metrics are obtained from Precision and Recall as follows.

\begin{equation}
    \text{Precision} = \frac{TP}{TP+FP}, \text{Recall} = \frac{TP}{TP+FN},
\end{equation}
\noindent
where TP, FP and FN are the true positive, false positive and false negative values obtained from the ground-truth $G$ and predicted binary mask $B$, both in the resolution of the input image domain $D_I$.

\paragraph{F-measure ($F_\beta$)} is given by:

\begin{equation}
    F_\beta = \frac{(1+\beta^2) \text{Precision}\times \text{Recall}}{\beta^2 \text{Precision}+\text{Recall}}.
\end{equation}

\paragraph{MAE} is given by:
 \begin{equation}
    \text{MAE} = \frac{1}{w\times h} \sum_{p\in D_I} |G(p) - B(p)|.
 \end{equation}
\noindent 

\subsection{Representative image selection}

We adopted a supervised approach to select a few representative images for $\mathcal{T}$ from the validation set $\mathcal{Z}_1\backslash \mathcal{T}$, since our focus is on the decoders, and $\mathcal{T}$ is paramount to obtain effective CNN models. Our approach is similar to the one proposed in~\cite{cerqueira2024interactive}. The method selects one image per time, by evaluating the current model on $\mathcal{Z}_1\backslash \mathcal{T}$. The selected images come from those not solved by the model trained on the current set $\mathcal{T}$ -- i.e., the ones from $\mathcal{Z}_1\backslash \mathcal{T}$ with the worst $F_{\beta}$ measure. However, differently from ~\cite{cerqueira2024interactive}, if the last image inserted in $\mathcal{T}$ does not produce a better model, we remove it, and try another one. 

The user decided when to stop the representative image selection process. In this work, the two users decided to stop the construction of $\mathcal{T}$ with three to four images, depending on the split. Given $\mathcal{T}=\{\}$ and $\mathcal{Z}_1$, this process can be described as follows (Algorithm \ref{alg:one}). 

\begin{algorithm}
    \caption{Representative image selection}\label{alg:one}
    \KwInput{$\mathcal{Z}_1, \mathcal{T} \leftarrow \{\}$}
    \KwOutput{$\mathcal{T} \neq \{\}$}
    Randomly select an image $z \in \mathcal{Z}_1$ and set $\mathcal{T} \leftarrow \mathcal{T} \cup\{z\}$;\\
    Set $x_{prev} \leftarrow 0$ and $z_{prev} \leftarrow z$;\\
    \While{User is not satisfied}{
        Train a FLIM CNN with an adaptive decoder on $\mathcal{T}$;\\
        Evaluate the model on $\mathcal{Z}_1\backslash \mathcal{T}$;\\
        Compute $x$ as the average $F_\beta$ on $\mathcal{Z}_1\backslash \mathcal{\mathcal{T}}$;\\
        Among the images with the lowest $F_{\beta}$, select a new image $z$  from $\mathcal{Z}_1\backslash \mathcal{T}$;\\
        \eIf{$x < x_{prev}$}{
            Set $\mathcal{T} \leftarrow \mathcal{T} \backslash \{z_{prev}\}$;\\
        }{
            Set $\mathcal{T} \leftarrow \mathcal{T} \cup\{z\}$;\\
            Set $x_{prev} \leftarrow x$ and $z_{prev} \leftarrow z$;
        }
    }
\end{algorithm}

For Parasites, one user selected FLIM$_{pb}$ as the adaptive decoder, and the other selected FLIM$_{lm}$. This variation prevents bias and ensures that adaptive decoders do not have an unfair advantage over fixed-weight decoders. For the BraTS dataset, one user selected FLIM$_{ts}$ as the adaptive decoder, and the other selected FLIM$_{lm}$.

\subsection{Post-processing}
\label{sec:postproc}

Post-processing differs as shown in \autoref{tab:post_processing}. Since the delineation loss in the lightweight models aims to create predicted masks with reasonable boundary delineation, we applied Otsu's threshold to the saliency maps of both datasets. We then removed false positives by an area filter. The area range was estimated from the data observation in $\mathcal{Z}_1\backslash \mathcal{T}$. This procedure was used for both datasets independently of the model, and it was enough for the FLIM CNNs in the case of the brain tumor dataset.

In the case of the Parasites, we also removed components connected to the image's frame (mostly false positives). We noticed that the predicted binary masks from FLIM CNNs, except \texorpdfstring{U-Net$_{\text{FLIM}}$}{} due to skip connection, are smaller than the object. We then applied morphological operations to create internal and external seeds and improve delineation by Dynamic Trees using the Lab color space of the original image~\citep{bragantini2018graph}. 

\begin{table}[!htbp]
\newcommand{\cellmultirow}[1]{\cellcolor{white}\multirow{-1}{*}{#1}}

\caption{Post-processing applied on the saliency maps of lightweight models, \texorpdfstring{U-Net$_{\text{FLIM}}$}{}, and the other FLIM models: OT is Otsu's Threshold; AF is Area Filtering using the range [minimum component area, maximum component area]; and DT is object delineation by Dynamic Trees~\citep{bragantini2018graph}}

\centering
\rowcolors{2}{gray!20}{white}
\begin{tabular}{cll}
    \hline \hline
     \hiderowcolors Dataset&Model& Post-processing\\\hline
     \showrowcolors
     & Lightweight & OT + AF[1000-9000]\\
   \cellmultirow{Parasites} & FLIM & OT + AF[1000-9000] + DT\\
    & \texorpdfstring{U-Net$_{\text{FLIM}}$}{} & OT + AF[1000-9000]\\
    \cellcolor{white}& Lightweight &OT + AF[100-20000]\\
    \cellmultirow{BraTS}& FLIM&OT + AF[100-20000]\\
    \cellcolor{white}& \texorpdfstring{U-Net$_{\text{FLIM}}$}{}&OT + AF[100-20000]\\
    \hline\hline
    \end{tabular}
\label{tab:post_processing}
\end{table}

\section{Results and discussion}\label{sec:results} 

This section presents the results of our experiments for each user, $A$ and $B$, using their respective training images and marker sets. The best FLIM network architectures (efficiency), quantitative results (effectiveness), and qualitative results are presented.

\subsection{The best FLIM architectures}\label{sec:best_flim_arch}

\begin{table}[!htbp]
\newcommand{\cellmultirow}[1]{\cellcolor{white}\multirow{-2}{*}{#1}}

\caption{Average number of parameters per user and model. FLIM networks that use a single decoder block are grouped for this measure.}

\centering
\begin{adjustbox}{width=0.48\textwidth}
\rowcolors{2}{gray!20}{white}
\begin{tabular}{llcc}
    \hline \hline
     \hiderowcolors\multicolumn{1}{c}{\multirow{2}{*}{Model}} & \multicolumn{1}{c}{\multirow{2}{*}{Users}} & \multicolumn{2}{|c}{\# Parameters} \\\cline{3-4}
     &\multicolumn{1}{c|}{}&Parasites&BraTS\\\hline
    \showrowcolors
    & A &1.33 (M) & 1.33 (M)\\
    \cellmultirow{SAMNet}& B &1.33 (M) & 1.33 (M)  \\

    & A & 3.26 (M) & 3.26 (M)\\
    \cellmultirow{MSCNet}& B & 3.26 (M) & 3.26 (M)\\

    &A & 3.27 (M) & 3.27 (M) \\
    \cellmultirow{MEANet}& B & 3.27 (M) & 3.27 (M)\\

    & A &$334.77\pm181.79$ (K)& $1846.88\pm839.77$ (K)\\
    \cellmultirow{FLIM}& B &$190.23\pm130.38$ (K)& $50.31\pm28.78$ (K)\\

    & A &$2.29\pm0.13$ (M)& $12.58\pm1.17$ (M)\\
    \cellmultirow{U-Net$_{\text{FLIM}}$}&B&$1.36\pm0.51$ (M)&$00.34\pm0.09$ (M)\\
    
    \hline\hline
    \end{tabular}
    \end{adjustbox}
\label{tab:architectures}
\end{table}

The numbers of blocks and kernels per block in the best FLIM encoder's architecture on $\mathcal{Z}_1\backslash \mathcal{T}$ change for each dataset, user, and split, affecting the complexity of the CNN architecture. User A, for instance, selected more markers for the BraTS dataset, increasing the number of parameters of his FLIM networks. \autoref{tab:architectures} shows the mean number of parameters in the best network architecture for each case. We are grouping all FLIM CNNs,  except  U-Net$_{\text{FLIM}}$ (due to its more complex decoder), for this efficiency measure with standard deviation. The lightweight models have a fixed number of parameters and they are hundreds of times more complex than FLIM CNNs, except U-Net$_{\text{FLIM}}$.

\begin{figure}[htb!]
\begin{center}
\includegraphics[width=\columnwidth]{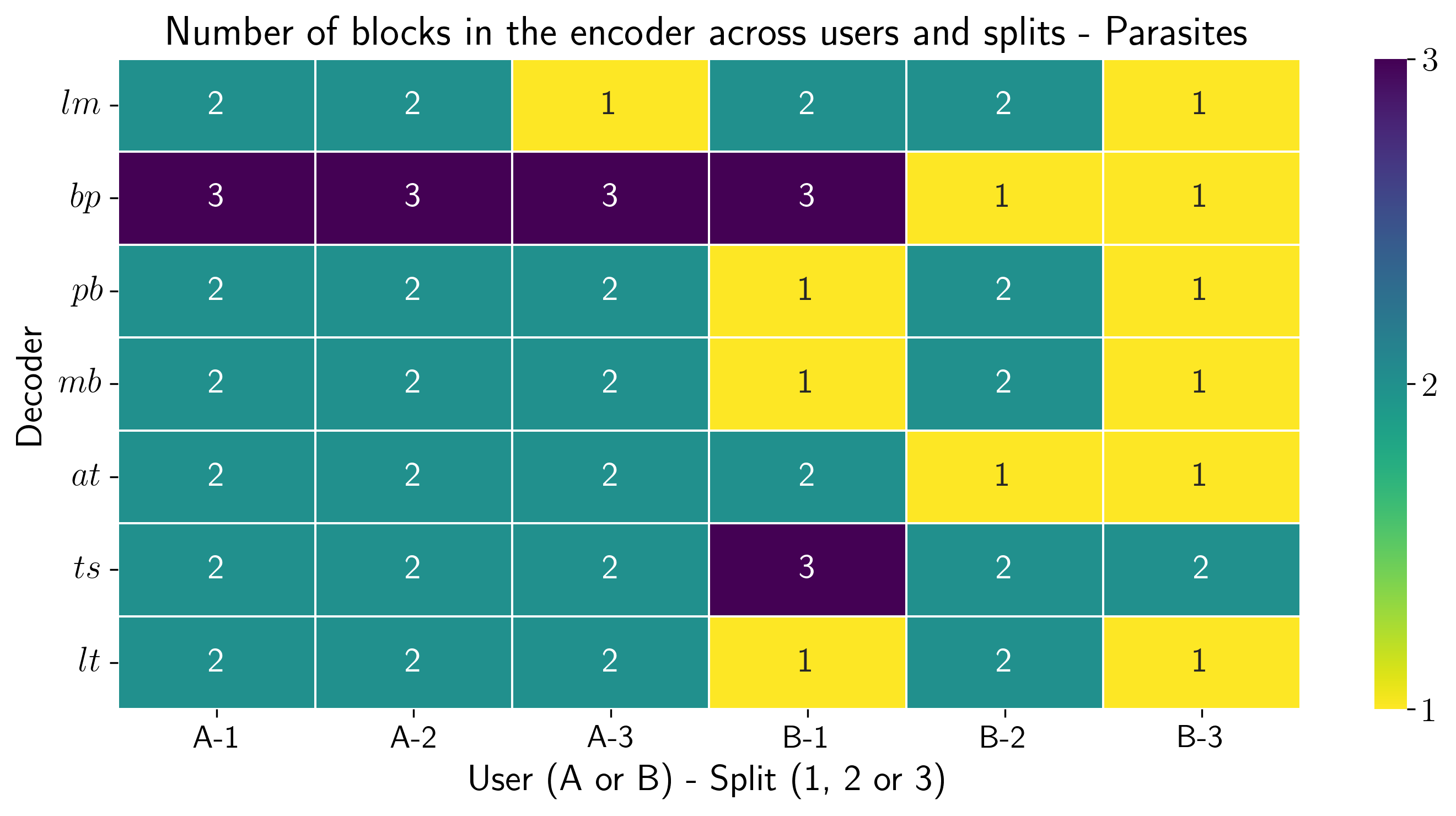}
\caption{Number of blocks in the FLIM encoder for the Parasites dataset.}
\label{fig:n_layers_schisto}
\end{center}
\end{figure}

\begin{figure}[htb!]
\begin{center}
\includegraphics[width=\columnwidth]{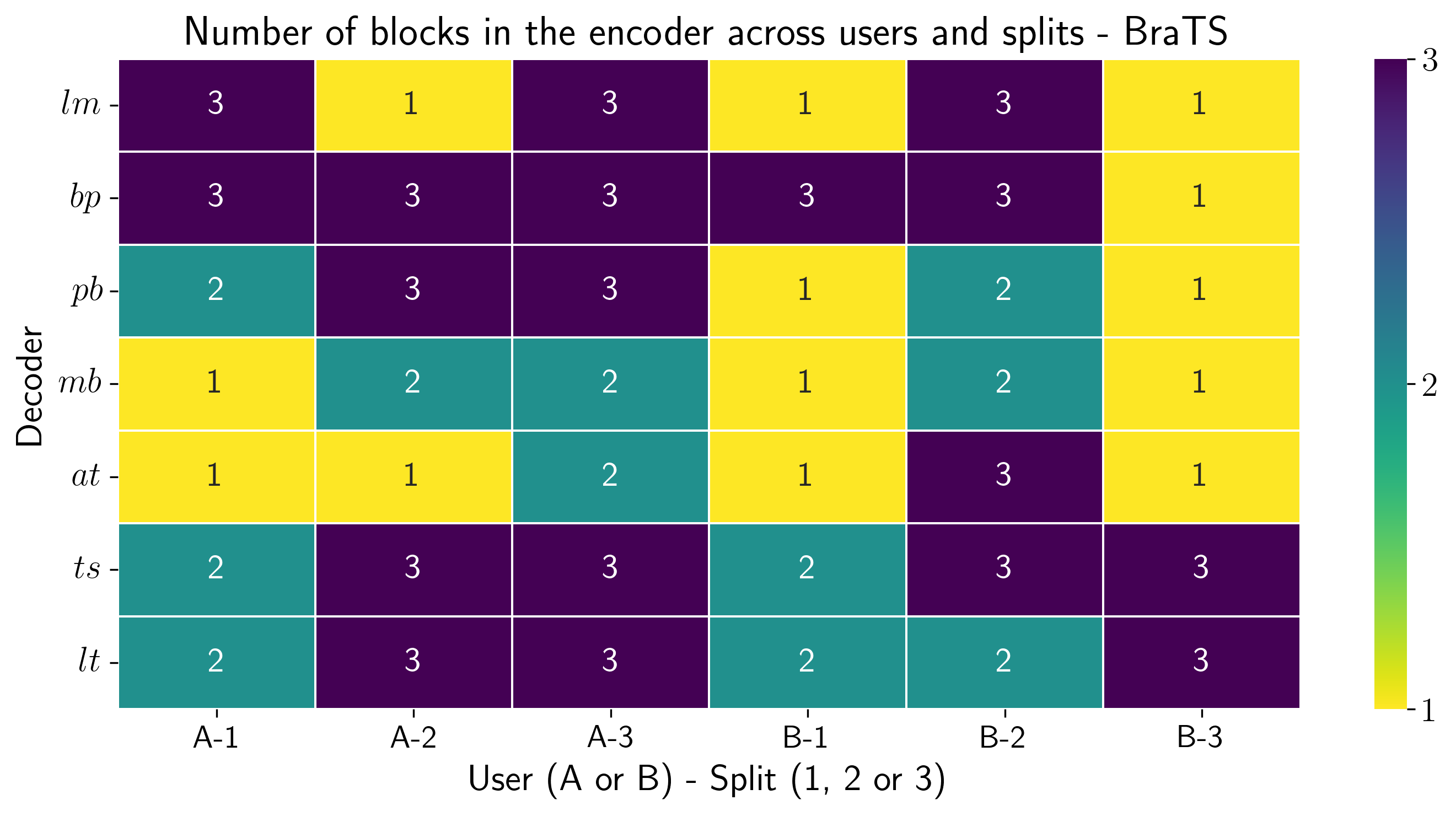}
\caption{Number of blocks in the FLIM encoder for the BraTS dataset.}
\label{fig:n_layers_brats}
\end{center}
\end{figure}

Figures~\ref{fig:n_layers_schisto} and~\ref{fig:n_layers_brats} show that the number of blocks in the best FLIM-CNN architecture varies a lot, depending on the dataset, split, decoder, and user. Note that an encoder with only two blocks was usually enough for Parasites. For the BraTS dataset, the best encoder depth has more variance. 

The encoder's depth also affects the decoders that combine the encoder's feature maps to create saliency maps. The deeper the encoder becomes,  the fewer false positives in the saliency map are observed, while more details of the object's border are lost (Figure~\ref{fig:delineation_result_blocks}). This behavior may improve object detection up to a certain depth but requires object delineation to improve effectiveness, which was the case with Parasites. 

\subsection{Quantitative results}

\autoref{tab:validation_set} shows the mean and standard deviation of MEA and $F_{beta}$ in the three splits per user on the validation sets $\mathcal{Z}_1\backslash \mathcal{T}$. The best results are highlighted in green, the second best in blue, and the worst results in red. For Parasites, FLIM$_{lm}$, FLIM$_{pb}$, and FLIM$_{mb}$ outperformed the other models in $MAE$ and $F_{\beta}$ for both users. According to both measures, these FLIM models are considerably more effective than the lightweight models, with SAMNet presenting the worst results. The good performance of FLIM$_{lm}$ indicates that the foreground and background activation channels created by the FLIM encoder consistently came from kernels with the same label, motivating the exploration of labeled markers for kernel estimation in future work.

Although FLIM$_{pb}$ and FLIM$_{mb}$, with the new decoders, present good MAE and $F_{\beta}$ for BraTS, FLIM$_{ts}$ and FLIM$_{lt}$ obtained slightly better results, while MSCNet and MEANet presented the worst MAE and $F_{\beta}$, respectively. Similar results can be observed in the test set $\mathcal{Z}_2$ of both datasets (\autoref{tab:test_set}), demonstrating the adaptive decoders' generalization and robustness to address SOD problems with distinct characteristics. 

\begin{table}[!t]
\newcommand{\cellmultirow}[1]{\cellcolor{white}\multirow{-2}{*}{#1}}

\caption{Mean and standard deviation of MEA and $F_{\beta}$ in the validation set. For each model and user, the best result is shown in \green{green}, the second best in \blue{blue}, and the worst in \red{red}. The central horizontal line separates FLIM CNNs with adaptive decoders from the baselines. Arrows $\uparrow$ and $\downarrow$ denote higher and lower values are better, respectively.}

\centering
\begin{adjustbox}{width=1.\columnwidth}
\rowcolors{2}{gray!20}{white}
\begin{tabularx}{1.5\columnwidth}{ll*{2}{c}*{2}{c}}
    \hline \hline
     \hiderowcolors \multicolumn{1}{X}{\multirow{2}{*}{Model}} & \multicolumn{1}{X}{\multirow{2}{*}{Users}} & \multicolumn{2}{|c}{Parasites} & \multicolumn{2}{|c}{BraTS}  \\ \cline{3-6}
    &\multicolumn{1}{c|}{}&MAE$\downarrow$&\multicolumn{1}{c|}{$F_\beta\uparrow$}&MAE$\downarrow$&$F_\beta\uparrow$ \\ \hline
    \showrowcolors
                                        &A&\red{0.038$\pm$0.036}&\red{0.428$\pm$0.234}            &0.140$\pm$0.072&0.219$\pm$0.078\\
\cellmultirow{SAMNet}               &B&\red{0.023$\pm$0.015}&\red{0.521$\pm$0.216}            &0.121$\pm$0.062&0.225$\pm$0.029\\

                                    &A&0.017$\pm$0.009&0.615$\pm$0.147                        &\red{0.197$\pm$0.047}&0.246$\pm$0.032\\
\cellmultirow{MSCNet}               &B&0.012$\pm$0.005&0.634$\pm$0.178                        &\red{0.161$\pm$0.020}&0.266$\pm$0.007\\
                        
                                    &A&0.023$\pm$0.017&0.538$\pm$0.120                        &0.049$\pm$0.003&\red{0.122$\pm$0.015}\\
\cellmultirow{MEANet}               &B&0.013$\pm$0.005&0.674$\pm$0.094                        &0.051$\pm$0.004&\red{0.131$\pm$0.022}\\

                                    &A&0.009$\pm$0.002&0.771$\pm$0.028                        &0.031$\pm$0.016&0.645$\pm$0.099\\
\cellmultirow{U-Net$_{\text{FLIM}}$} &B&0.011$\pm$0.004&0.710$\pm$0.087                        &0.049$\pm$0.037&0.630$\pm$0.097\\

                                    &A&\green{0.005$\pm$0.001}&\green{0.860$\pm$0.017}        &0.018$\pm$0.001&0.659$\pm$0.021\\
\cellmultirow{FLIM$_{lm}$}          &B&\green{0.005$\pm$0.001}&0.843$\pm$0.025                 &0.024$\pm$0.007&0.691$\pm$0.012\\

                                    &A&0.007$\pm$0.000&0.770$\pm$0.030                        &0.019$\pm$0.001&0.707$\pm$0.005\\
\cellmultirow{FLIM$_{bp}$}          &B&0.008$\pm$0.001&0.737$\pm$0.025                        &0.023$\pm$0.007&0.693$\pm$0.014\\\hline

                                    &A&\blue{0.006$\pm$0.001}&\blue{0.857$\pm$0.012}          &0.019$\pm$0.000&0.703$\pm$0.011\\ 
\cellmultirow{FLIM$_{pb}$}          &B&\blue{0.006$\pm$0.001}&\green{0.847$\pm$0.015}         &0.022$\pm$0.003&0.691$\pm$0.008\\

                                    &A&0.006$\pm$0.002&0.843$\pm$0.023                        &0.021$\pm$0.002&0.702$\pm$0.006\\
\cellmultirow{FLIM$_{mb}$}          &B&0.006$\pm$0.001&\blue{0.843$\pm$0.021}                        &0.025$\pm$0.007&0.694$\pm$0.017\\

                                    &A&0.011$\pm$0.003&0.740$\pm$0.062                        &0.022$\pm$0.003&0.679$\pm$0.013\\
\cellmultirow{FLIM$_{at}$}          &B&0.014$\pm$0.006&0.660$\pm$0.108                        &0.024$\pm$0.006&0.694$\pm$0.017\\

                                    &A&0.010$\pm$0.000&0.747$\pm$0.015                        &\green{0.017$\pm$0.001}&\blue{0.709$\pm$0.014}\\
\cellmultirow{FLIM$_{ts}$}          &B&0.013$\pm$0.004&0.687$\pm$0.085                        &\green{0.020$\pm$0.004}&\green{0.697$\pm$0.037}\\

                                    &A&0.007$\pm$0.001&0.810$\pm$0.010                        &\blue{0.018$\pm$0.001}&\green{0.721$\pm$0.004}\\
\cellmultirow{FLIM$_{lt}$}          &B&0.009$\pm$0.004&0.760$\pm$0.089                        &\blue{0.021$\pm$0.005}&\blue{0.697$\pm$0.044}\\
    \hline\hline
    \end{tabularx}
\end{adjustbox}
\label{tab:validation_set}
\end{table}

\begin{table}[!htbp]
\newcommand{\cellmultirow}[1]{\cellcolor{white}\multirow{-2}{*}{#1}}

\caption{Mean and standard deviation of MEA and $F_{\beta}$ in the test set. For each model and user, the best result is shown in \green{green}, the second best in \blue{blue}, and the worst in \red{red}. The central horizontal line separates FLIM CNNs with adaptive decoders from the baselines. Arrows $\uparrow$ and $\downarrow$ denote higher and lower values are better, respectively.}
\centering
\begin{adjustbox}{width=\columnwidth}
\rowcolors{2}{gray!20}{white}
\begin{tabularx}{1.5\columnwidth}{ll*{2}{c}*{2}{c}}
    \hline \hline
     \hiderowcolors \multicolumn{1}{X}{\multirow{2}{*}{Model}} & \multicolumn{1}{X}{\multirow{2}{*}{Users}} & \multicolumn{2}{|c}{Parasites} & \multicolumn{2}{|c}{BraTS}  \\ \cline{3-6}
    &\multicolumn{1}{c|}{}&MAE$\downarrow$&\multicolumn{1}{c|}{$F_\beta\uparrow$}&MAE$\downarrow$&$F_\beta\uparrow$ \\ \hline
    \showrowcolors
            &A&\red{0.038$\pm$0.036}&\red{0.422$\pm$0.241}            &0.142$\pm$0.068&0.215$\pm$0.084\\
\cellmultirow{SAMNet}               &B&\red{0.024$\pm$0.014}&\red{0.508$\pm$0.194}            &0.123$\pm$0.063&0.221$\pm$0.038\\

                                    &A&0.016$\pm$0.009&0.604$\pm$0.125                        &\red{0.197$\pm$0.047}&0.248$\pm$0.028\\
\cellmultirow{MSCNet}               &B&0.012$\pm$0.005&0.645$\pm$0.163                        &\red{0.161$\pm$0.020}&0.268$\pm$0.003\\
                        
                                    &A&0.023$\pm$0.018&0.538$\pm$0.137                        &0.050$\pm$0.004&\red{0.124$\pm$0.010}\\
\cellmultirow{MEANet}               &B&0.012$\pm$0.004&0.681$\pm$0.085                        &0.052$\pm$0.003&\red{0.126$\pm$0.019}\\

                                    &A&0.008$\pm$0.001&0.777$\pm$0.025                        &0.032$\pm$0.017&0.650$\pm$0.104\\
\cellmultirow{U-Net$_{\text{FLIM}}$} &B&0.011$\pm$0.004&0.706$\pm$0.063                        &0.049$\pm$0.038&0.636$\pm$0.103\\

                                    &A&\green{0.005$\pm$0.000}&\blue{0.857$\pm$0.015}         &0.018$\pm$0.001&0.678$\pm$0.040\\
\cellmultirow{FLIM$_{lm}$}          &B&\green{0.005$\pm$0.000}&0.843$\pm$0.015                &0.023$\pm$0.007&0.700$\pm$0.013\\

                                    &A&0.007$\pm$0.001&0.757$\pm$0.012                        &0.018$\pm$0.001&0.717$\pm$0.005\\
\cellmultirow{FLIM$_{bp}$}          &B&0.008$\pm$0.001&0.733$\pm$0.025                        &0.023$\pm$0.007&0.704$\pm$0.016\\\hline

                                    &A&\blue{0.006$\pm$0.000}&\green{0.857$\pm$0.012}         &0.019$\pm$0.001&0.711$\pm$0.010\\ 
\cellmultirow{FLIM$_{pb}$}          &B&\blue{0.006$\pm$0.001}&\green{0.850$\pm$0.010}         &\blue{0.021$\pm$0.003}&0.701$\pm$0.009\\

                                    &A&0.006$\pm$0.001&0.847$\pm$0.012                        &0.021$\pm$0.002&0.712$\pm$0.007\\
\cellmultirow{FLIM$_{mb}$}          &B&0.006$\pm$0.001&\blue{0.850$\pm$0.010}                 &0.025$\pm$0.007&0.702$\pm$0.019\\

                                    &A&0.011$\pm$0.002&0.733$\pm$0.042                        &0.021$\pm$0.003&0.688$\pm$0.012\\
\cellmultirow{FLIM$_{at}$}          &B&0.015$\pm$0.006&0.647$\pm$0.101                        &0.024$\pm$0.006&0.703$\pm$0.018\\

                                    &A&0.010$\pm$0.001&0.747$\pm$0.015                        &\green{0.017$\pm$0.001}&\blue{0.720$\pm$0.014}\\
\cellmultirow{FLIM$_{ts}$}          &B&0.013$\pm$0.003&0.647$\pm$0.057                        &\green{0.019$\pm$0.004}&\green{0.706$\pm$0.039}\\

                                    &A&0.007$\pm$0.001&0.803$\pm$0.021                        &\blue{0.017$\pm$0.002}&\green{0.731$\pm$0.005}\\
\cellmultirow{FLIM$_{lt}$}          &B&0.009$\pm$0.005&0.753$\pm$0.086                        &0.021$\pm$0.005&\blue{0.704$\pm$0.048}\\
\hline\hline
    \end{tabularx}
    \label{tab:test_set}
\end{adjustbox}
\end{table}

Since the models were trained with very few images (three or four), a question may arise about the performance of the backpropagation-based baselines when a reasonable number of annotated images is available. For that, we trained the more complex baselines based on backpropagation on the entire set $\mathcal{Z}_1$ and tested them on $\mathcal{Z}_2$. For U-Net$_\text{FLIM}$, we use the encoder unfrozen to fine-tune the FLIM encoder with the higher number of annotated images in $\mathcal{Z}_1$. 

\autoref{tab:experiments_more_images} shows that SAMNet and MSCNet could improve their results on Parasites, achieving superior MAE and $F_\beta$ to the others. However, they could not perform well on BraTS, and MEANnet could not perform well in both datasets, indicating that the number of annotated images in $\mathcal{Z}_1$ was still insufficient. The Parasites dataset contains most images in $\mathcal{Z}_1$ with only impurities, and this class imbalance affected MEANnet but not SAMNet and MSCNet. U-Net$_\text{FLIM}$ improved $F_\beta$ and MAE in both datasets, producing superior results to the others in BraTS. 

The results in \autoref{tab:experiments_more_images} reveal the robustness of FLIM SOD networks with adaptive decoders trained from markers (weak annotation) on very few representative images. They also demonstrate that the small set $\mathcal{T}$ is representative to train effective FLIM SOD networks.

\begin{table}[!t]
\newcommand{\cellmultirow}[1]{\cellcolor{white}\multirow{-2}{*}{#1}}

\caption{Mean and standard deviation of MEA and $F_{\beta}$ on the test set $\mathcal{Z}_2$ of the baseline backpropagation-based models, when trained with the entire set $\mathcal{Z}_1$. The results of the FLIM networks are presented with the encoder trained by each user, A and B. Arrows $\uparrow$ and $\downarrow$ denote higher and lower values are better, respectively.}

\centering
\begin{adjustbox}{width=1.\columnwidth}
\rowcolors{2}{gray!20}{white}
\begin{tabularx}{1.5\columnwidth}{l*{2}{c}*{2}{c}}
    \hline \hline
     \hiderowcolors \multicolumn{1}{X}{\multirow{2}{*}{Model}}& \multicolumn{2}{|c}{Parasites} & \multicolumn{2}{|c}{BraTS}  \\ \cline{2-5}
    \multicolumn{1}{c|}{}&MAE$\downarrow$&\multicolumn{1}{c|}{$F_\beta\uparrow$}&MAE$\downarrow$&$F_\beta\uparrow$ \\ \hline
    \showrowcolors
    SAMNet&0.002$\pm$0.001&0.939$\pm$0.011  &0.081$\pm$0.018&0.257$\pm$0.047\\
    MSCNet&0.002$\pm$0.000&0.933$\pm$0.011  &0.057$\pm$0.001&0.360$\pm$0.003\\
    MEANet&0.042$\pm$0.016&0.529$\pm$0.023  &0.044$\pm$0.001&0.257$\pm$0.006\\         
    U-Net$_{\text{FLIM}}$ (A)&0.004$\pm$0.001&0.815$\pm$0.041&0.009$\pm$0.000&0.855$\pm$0.008\\
    U-Net$_{\text{FLIM}}$ (B)&0.004$\pm$0.001&0.865$\pm$0.008&0.009$\pm$0.000&0.852$\pm$0.005\\
    \hline\hline
    \end{tabularx}
\end{adjustbox}
\label{tab:experiments_more_images}
\end{table}

\subsection{Qualitative results}

\begin{figure*}[htb!]
\begin{center}
\newcommand\sizefig{0.121\textwidth}
\centering
\makebox[\textwidth][c]{%
\renewcommand{\arraystretch}{1}
\begin{tabular}{*{8}{c@{\hskip 0.5pt}}c}
\includegraphics[width=\sizefig]{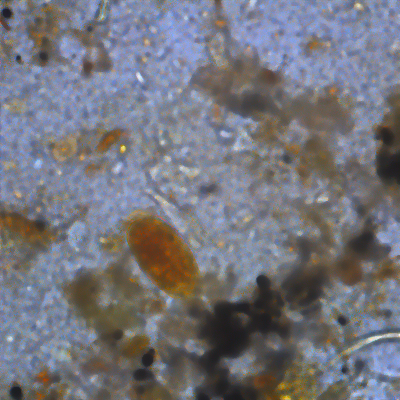}&
\includegraphics[width=\sizefig]{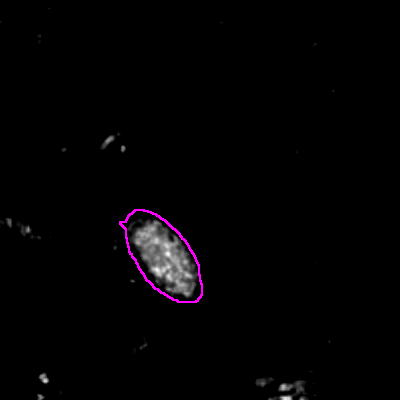}&
\includegraphics[width=\sizefig]{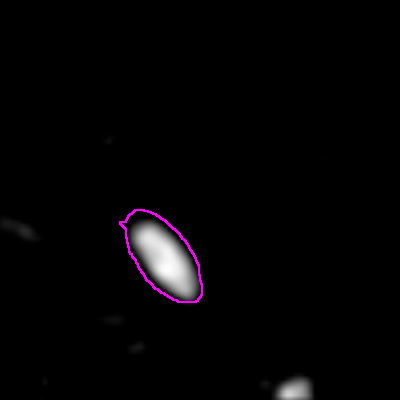}&
\includegraphics[width=\sizefig]{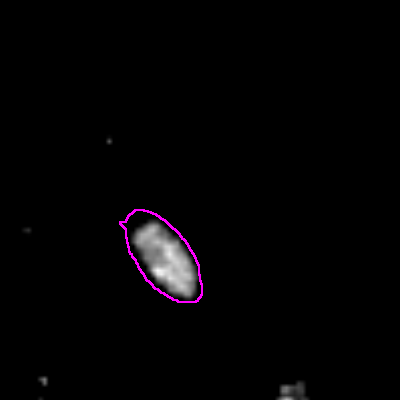}&
\includegraphics[width=\sizefig]{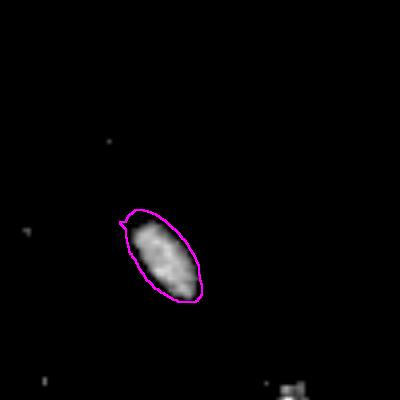}&
\includegraphics[width=\sizefig]{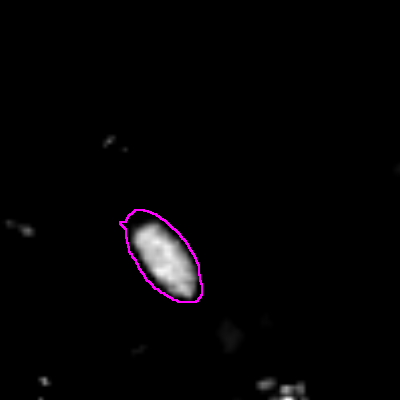}&
\includegraphics[width=\sizefig]{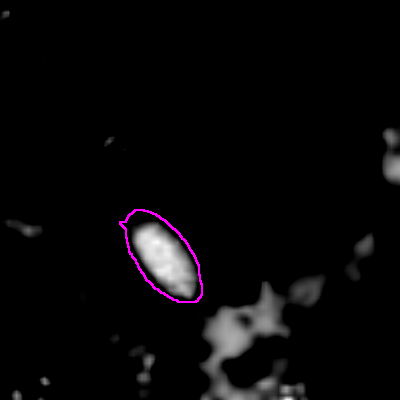}&
\includegraphics[width=\sizefig]{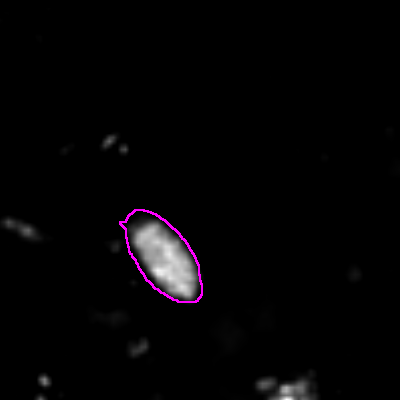}\\
\includegraphics[width=\sizefig]{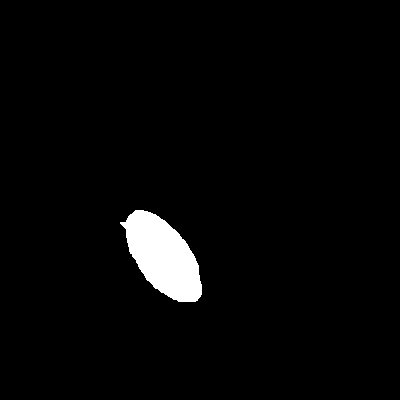}&
\includegraphics[width=\sizefig]{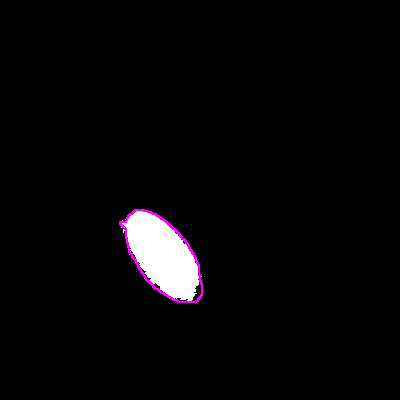}&
\includegraphics[width=\sizefig]{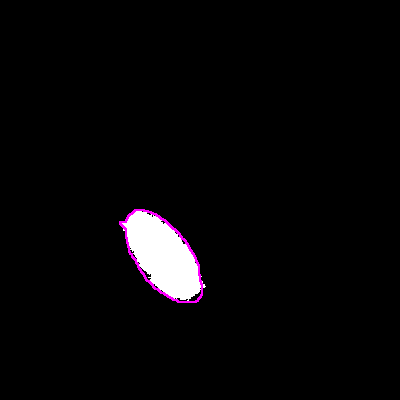}&
\includegraphics[width=\sizefig]{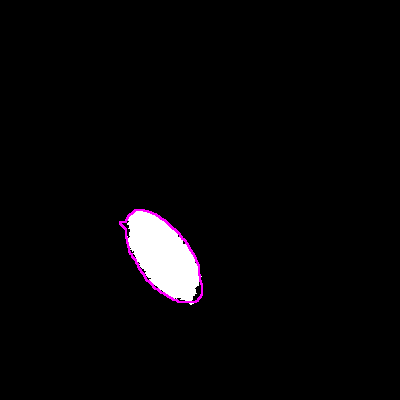}&
\includegraphics[width=\sizefig]{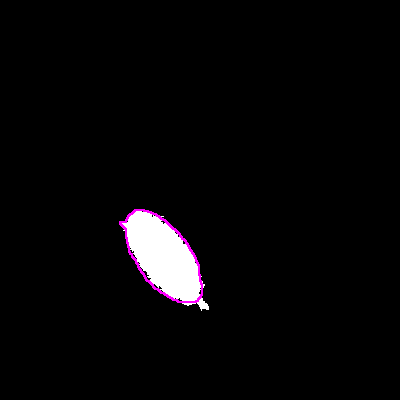}&
\includegraphics[width=\sizefig]{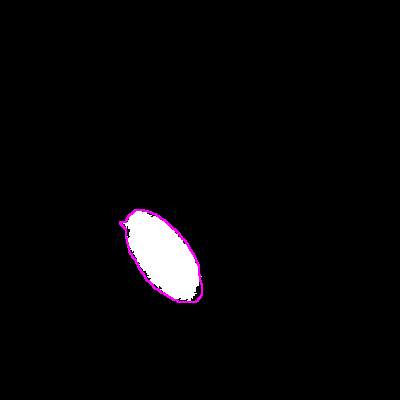}&
\includegraphics[width=\sizefig]{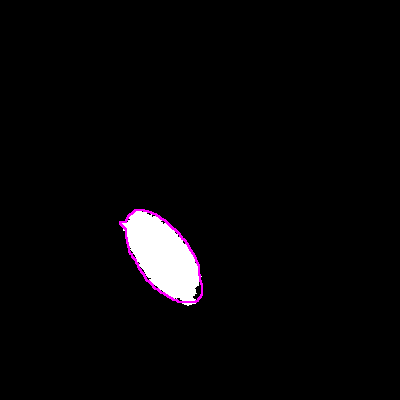}&
\includegraphics[width=\sizefig]{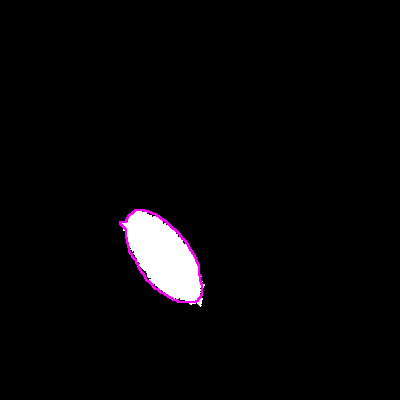}\\
\footnotesize(a) Original image & (b) $lm$ & (c) $bp$ & (d) $pb$ & (e) $mb$ & (f) $at$ & (g) $ts$ & (h) $ts$\\
\end{tabular}
}
\caption{
(a) Original image (above) and ground-truth mask (below) of an example from the Parasites. (b-g) show the improvement in salient object detection when comparing the adaptive decoders' output (above) with the results of post-processing (below) for each network.}
    \label{fig:delineation_result}
\end{center}
\end{figure*}

\begin{figure*}[hbt!]
\newcommand\sizefig{0.165\textwidth}
\newcommand\figspace{\hspace{0.1cm}}
\centering
\makebox[\textwidth][c]{%
\renewcommand{\arraystretch}{1}
\begin{tabular}{*{6}{c@{\hskip 0.5pt}}c}
\includegraphics[width=\sizefig]{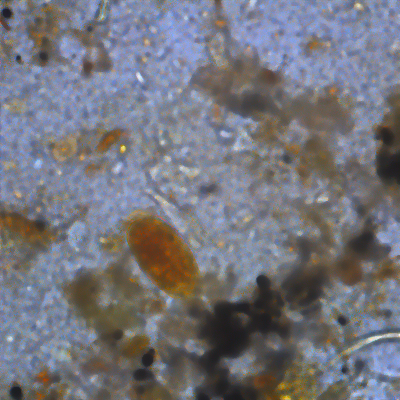}  &
\includegraphics[width=\sizefig]{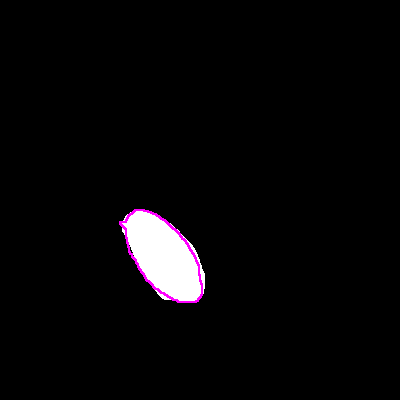}  &
\includegraphics[width=\sizefig]{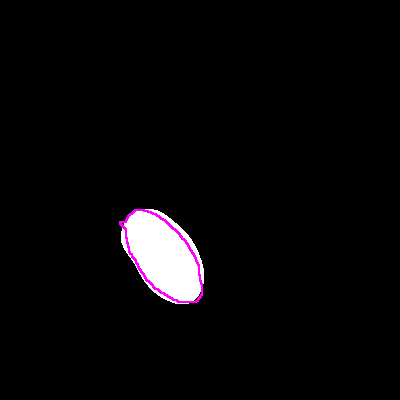}  &
\includegraphics[width=\sizefig]{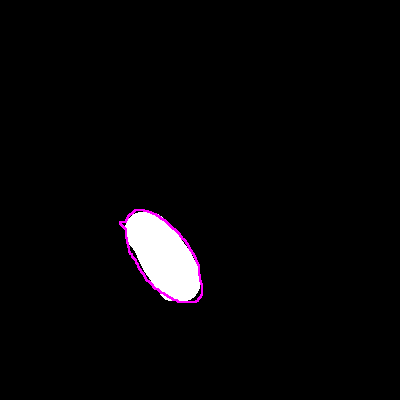}  &
\includegraphics[width=\sizefig]{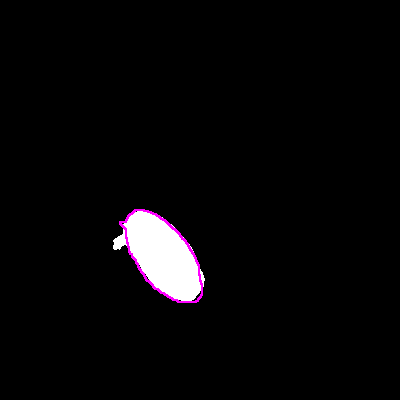}    &
\includegraphics[width=\sizefig]{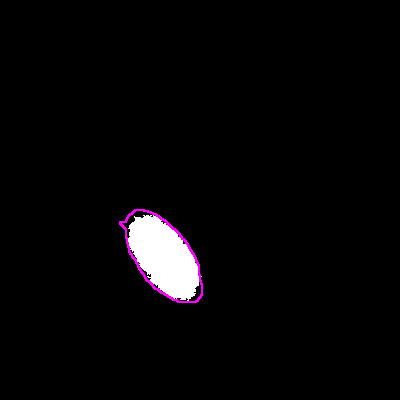}      & \vspace{-3pt}\\
\includegraphics[width=\sizefig]{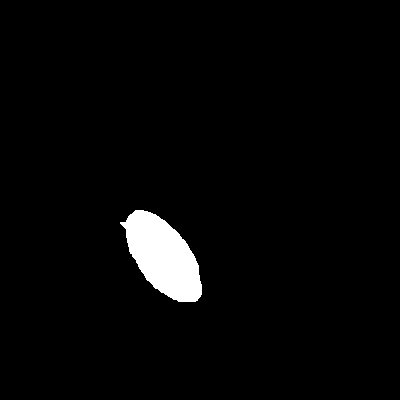}        &
\includegraphics[width=\sizefig]{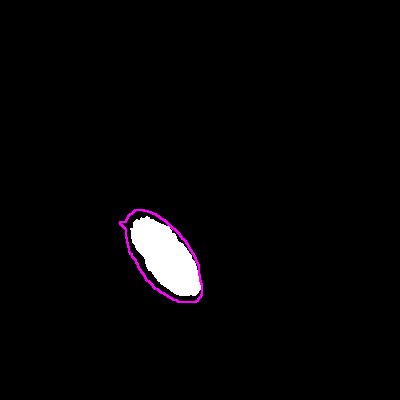}  &
\includegraphics[width=\sizefig]{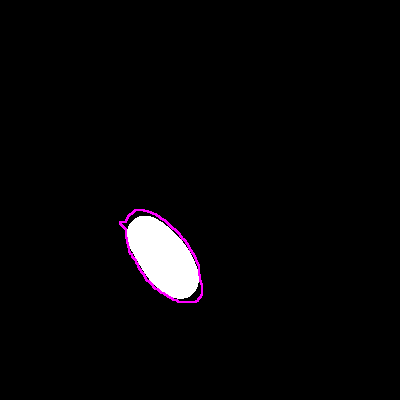}  &
\includegraphics[width=\sizefig]{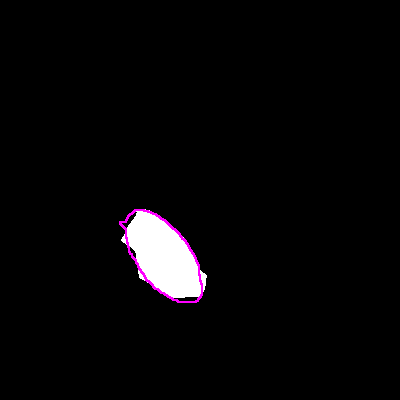}  &
\includegraphics[width=\sizefig]{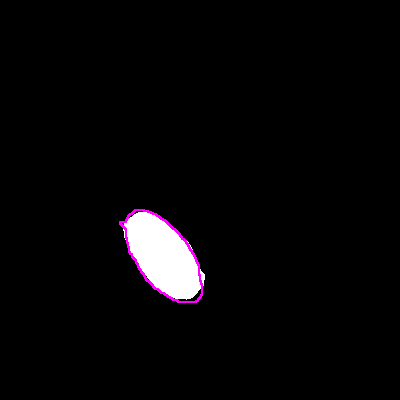}    &
\includegraphics[width=\sizefig]{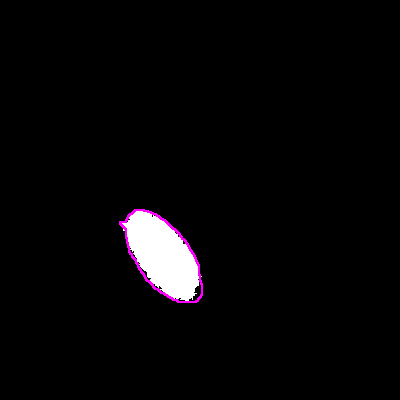}      & \\
(a) Original image & (b) SAMNet & (c) MSCNet & (d) MEANet & (e) U-Net$_{\text{FLIM}}$ & (f) $lm$\\
\includegraphics[width=\sizefig]{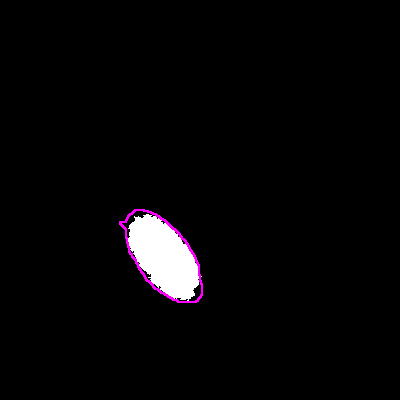}       &
\includegraphics[width=\sizefig]{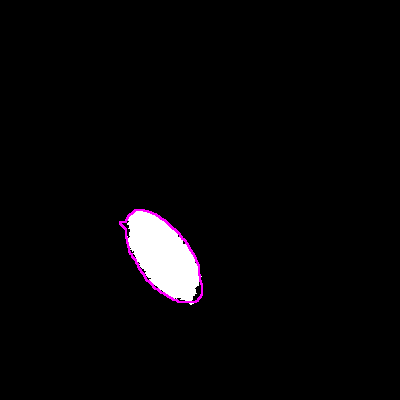}       &
\includegraphics[width=\sizefig]{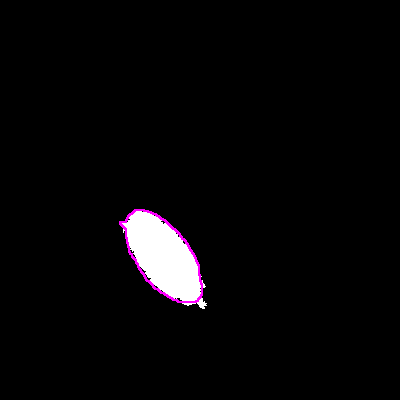}       &
\includegraphics[width=\sizefig]{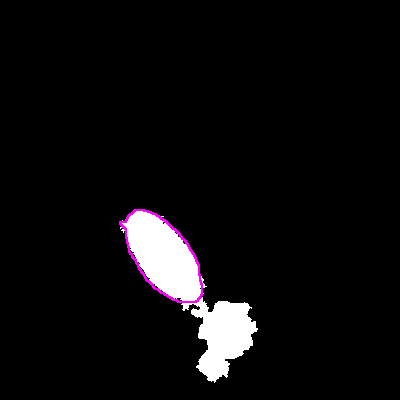}       &
\includegraphics[width=\sizefig]{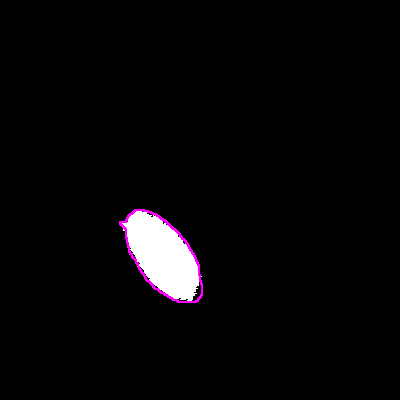}       &
\includegraphics[width=\sizefig]{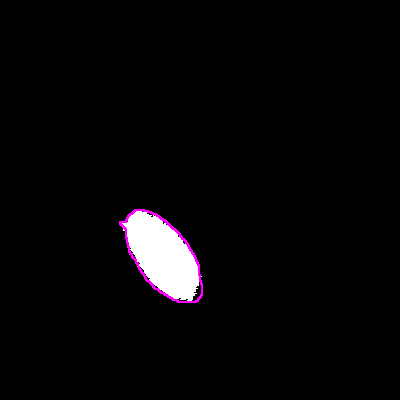}       & \vspace{-3pt}\\
\includegraphics[width=\sizefig]{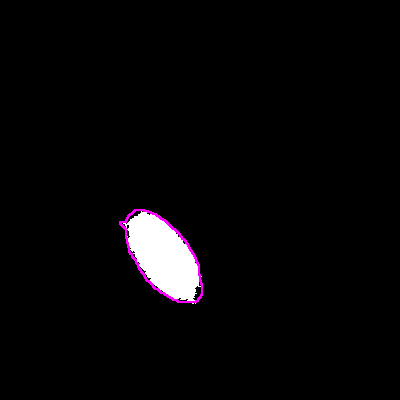}       &
\includegraphics[width=\sizefig]{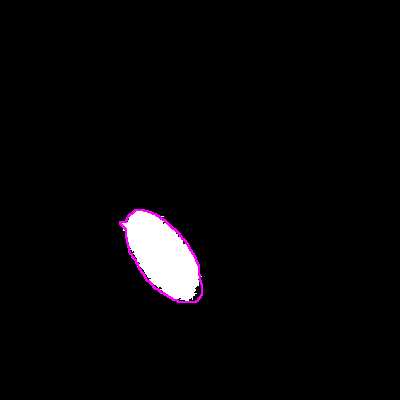}       &
\includegraphics[width=\sizefig]{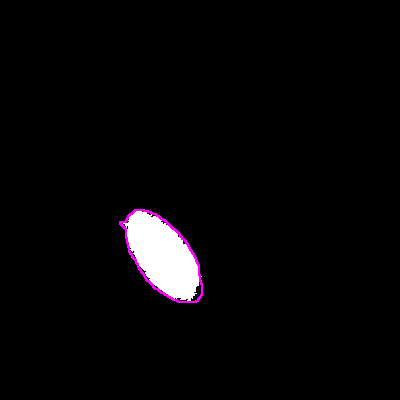}       &
\includegraphics[width=\sizefig]{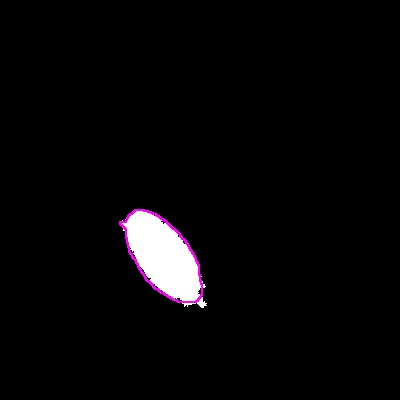}       &
\includegraphics[width=\sizefig]{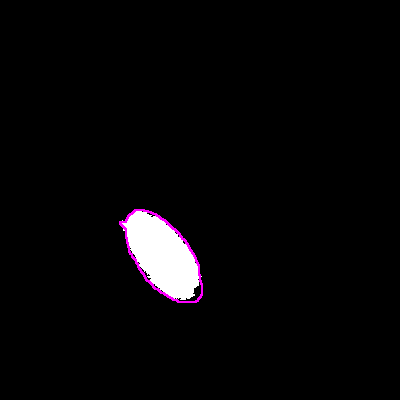}       &
\includegraphics[width=\sizefig]{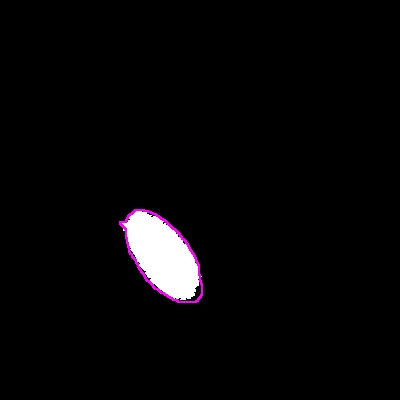}       & \\
(g) $bp$ & (h) $pb$ & (i) $mb$ & (j) $at$ & (k) $ts$ & (l) $lt$
\end{tabular}
}
\caption{(a) Original image (above) and ground-truth mask (below) of an example from the Parasites. (b-g) show the qualitative results of the baseline networks trained by users A (above) and B (below). (h-l) show the qualitative results of the networks with adaptive decoders trained by users A (above) and B (below).}
\label{fig:qualitative_results1}
\end{figure*}

\autoref{fig:delineation_result} illustrates the improvement in salient object detection for an example from Parasites when post-processing is applied to the adaptive decoder's output. Note that Dynamic Trees is responsible for improving object delineation. \autoref{fig:delineation_result}a shows the original image (above) and its ground-truth mask (below). The superior row shows the saliency maps at the adaptive decoder's output for each network, and the inferior row shows the respective results from post-processing for each network (Figures~\ref{fig:delineation_result}b-\ref{fig:delineation_result}g). The border of the ground-truth mask is shown in color to illustrate accuracy. The presented results are good, but errors in delineation may occur when both the parasite egg and connected impurity components are enhanced in the saliency map (e.g.,~\autoref{fig:qualitative_results1}j, above).

\begin{figure*}[hbt!]
\newcommand\sizefig{0.165\textwidth}
\newcommand\figspace{\hspace{0.1cm}}
\centering
\makebox[\textwidth][c]{%
\renewcommand{\arraystretch}{1}
\begin{tabular}{*{6}{c@{\hskip 0.5pt}}c}
\includegraphics[width=\sizefig]{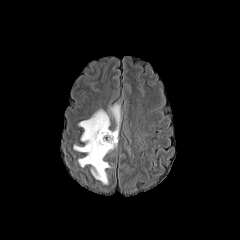}  &
\includegraphics[width=\sizefig]{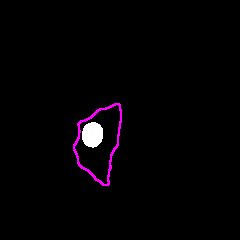}  &
\includegraphics[width=\sizefig]{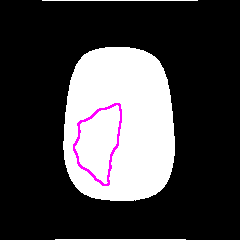}  &
\includegraphics[width=\sizefig]{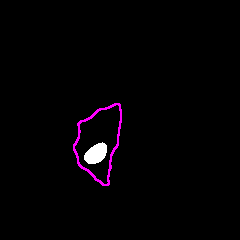}  &
\includegraphics[width=\sizefig]{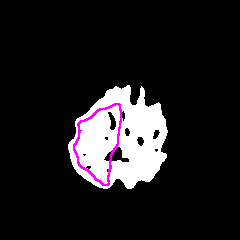}    &
\includegraphics[width=\sizefig]{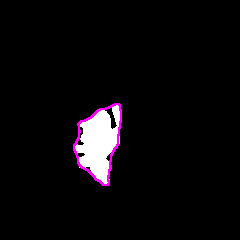}      & \vspace{-3pt}\\
\includegraphics[width=\sizefig]{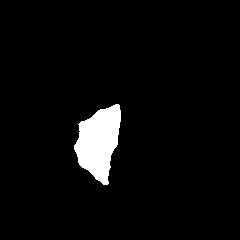}        &
\includegraphics[width=\sizefig]{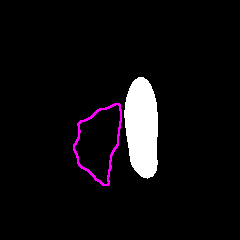}  &
\includegraphics[width=\sizefig]{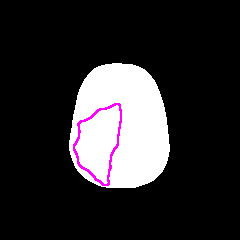}  &
\includegraphics[width=\sizefig]{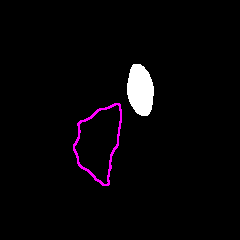}  &
\includegraphics[width=\sizefig]{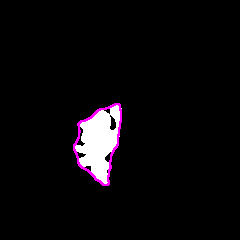}    &
\includegraphics[width=\sizefig]{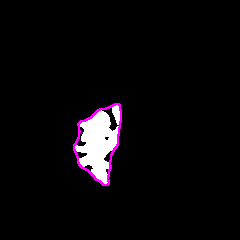}      & \\
(a) Original image & (b) SAMNet & (c) MSCNet & (d) MEANet & (e) U-Net$_{\text{FLIM}}$ & (f) $lm$\\
\includegraphics[width=\sizefig]{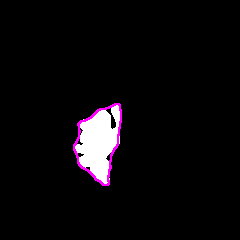}       &
\includegraphics[width=\sizefig]{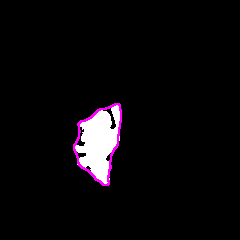}       &
\includegraphics[width=\sizefig]{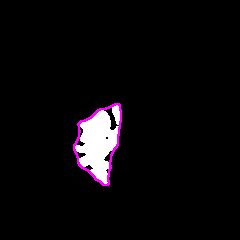}       &
\includegraphics[width=\sizefig]{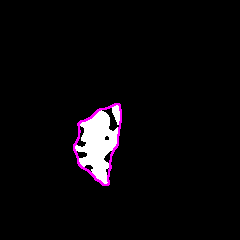}       &
\includegraphics[width=\sizefig]{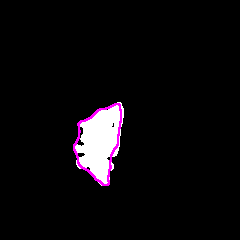}       &
\includegraphics[width=\sizefig]{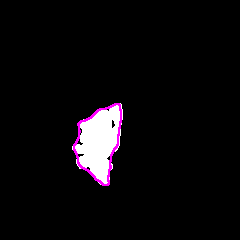}       & \vspace{-3pt}\\
\includegraphics[width=\sizefig]{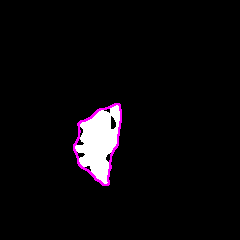}       &
\includegraphics[width=\sizefig]{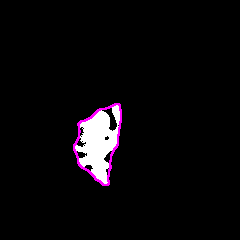}       &
\includegraphics[width=\sizefig]{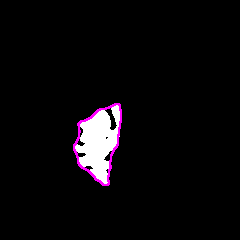}       &
\includegraphics[width=\sizefig]{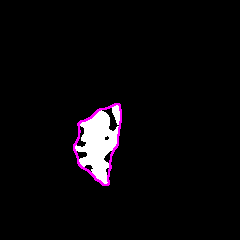}       &
\includegraphics[width=\sizefig]{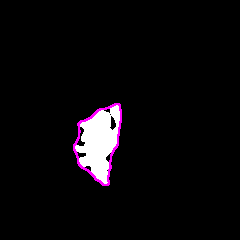}       &
\includegraphics[width=\sizefig]{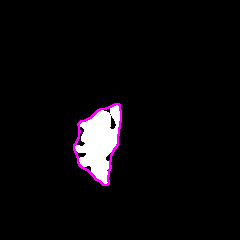}       & \\
(g) $bp$ & (h) $pb$ & (i) $mb$ & (j) $at$ & (k) $ts$ & (l) $lt$
\end{tabular}
}
\caption{(a) Original image (above) and ground-truth mask (below) of an example from the BraTS dataset. (b-g) show the qualitative results of the baseline networks trained by users A (above) and B (below). (h-l) show the qualitative results of the networks with adaptive decoders trained by users A (above) and B (below).}
\label{fig:qualitative_results2}
\end{figure*}

\begin{figure*}[hbt!]
\newcommand\sizefig{0.195\textwidth}
\centering
\makebox[\textwidth][c]{%
\renewcommand{\arraystretch}{1}
\begin{tabular}{*{6}{c@{\hskip 0.5pt}}c}
\includegraphics[width=\sizefig]{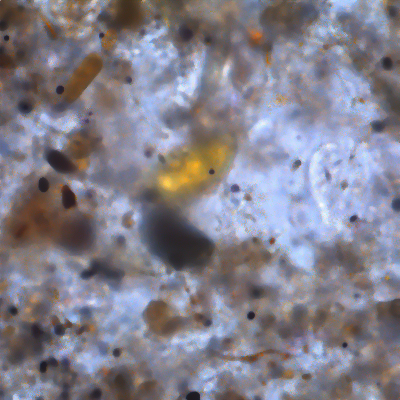}  &
\includegraphics[width=\sizefig]{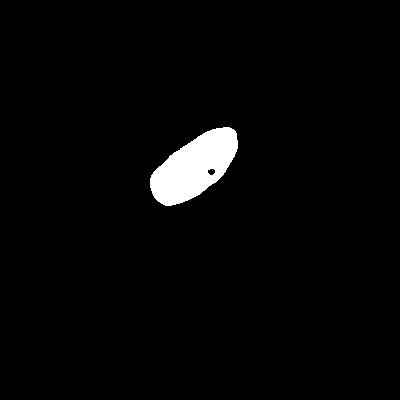}  &
\includegraphics[width=\sizefig]{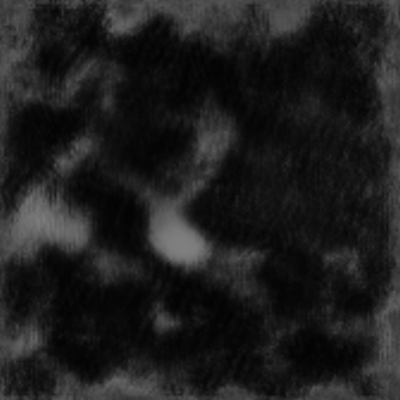}  &
\includegraphics[width=\sizefig]{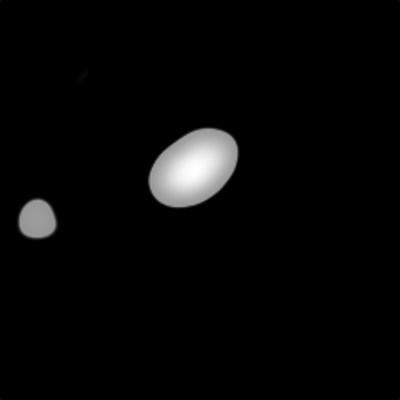}  &
\includegraphics[width=\sizefig]{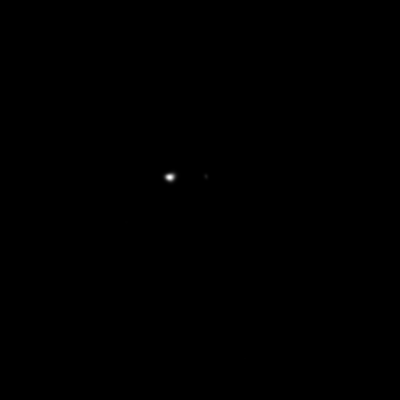}   \\
(a) Original image & (b) Ground truth & (c) SAMNet & (d) MSCNet & (e) MEANet\\
\end{tabular}
}
\caption{This example illustrates typical convergence errors of the lightweight models in a training image.}
\label{fig:lightweights_fail}
\end{figure*}

Figures~\ref{fig:qualitative_results1} and~\ref{fig:qualitative_results2} show qualitative results of the baselines and FLIM networks with adaptive decoders trained by users A (above row) and B (below row) for examples from Parasites and BraTS, respectively. These examples illustrate the better delineation observed in the quantitative results for Parasites compared to BraTS. They also demonstrate the type of errors observed with the lightweight models. Indeed, such models presented convergence errors in some training images~\autoref{fig:lightweights_fail}.

\section{Conclusion}\label{sec:conclusion}

This work presented five adaptive decoders for FLIM networks, evaluating them in two SOD tasks: (a) parasite egg detection in optical microscopy images (dataset Parasites) and (b) brain tumor detection in magnetic resonance slices (dataset BraTS). The adaptive decoders FLIM$_{ts}$, FLIM$_{at}$, and FLIM$_{lt}$ were previously presented in~\cite{soares2024adaptive}, while FLIM$_{pb}$ and FLIM$_{mb}$ were introduced here. Unlike the previous ones,  FLIM$_{pb}$ and FLIM$_{mb}$ estimate adaptive weights per pixel, being more complex than a simple point-wise convolution followed by activation.

The work demonstrated that one can create efficient and effective networks for SOD from very few representative images (three-four), by combining a FLIM encoder with an adaptive decoder. It described how representative images were selected, and compared FLIM networks with adaptive decoders against three pre-trained lightweight models, SAMNet, MSCNet, and MEANet, two FLIM networks with decoders trained by backpropagation, U-Net$_{FLIM}$ and FLIM$_{bp}$, and one network, FLIM$_{lm}$, whose marker labels define the decoder's weights. FLIM$_{pb}$, FLIM$_{mb}$, and FLIM$_{lm}$ outperformed the others in Parasites, while FLIM$_{ts}$ and FLIM$_{lt}$ were the best models in BraTS. Indeed, compared to the lightweight models, all FLIM networks, including the baseline ones, presented better results in both datasets. 
Even when trained on the full dataset $\mathcal{Z}_1$, most lightweight models --except for SAMNet and MSCNet on Parasites-- failed to improve or match the performance of FLIM networks.

To conclude, the lightweight models likely require considerably more annotated samples to achieve meaningful performance gains. In contrast, FLIM networks equipped with adaptive decoders offer a superior alternative, particularly in resource-constrained settings.

The post-processing object delineation worked effectively for Parasites, but not for BraTS, likely due to the choice of delineation algorithm--a subject warranting further investigation. Additionally, combining saliency maps from different encoder blocks merits deeper exploration, as early blocks often capture object boundaries more accurately, while later blocks reduce false positives by better localizing the object. Future work will also focus on developing new adaptive decoders for broader applications.

\section*{contributions}
GJS, SJFG, LN, and AXF conceptualized this study. JFG contributed to data curation. GJS and MAC performed the experiments. AXF supervised GJS. GJS wrote the original draft. AXF, LN, and SJFG reviewed and edited the text.  All authors read and approved the final manuscript.

\section*{acknowledgements}
The authors acknowledge grants from Quinto Andar, FAPESP (2023/14427-8, 2023/09210-0 and 2013/07375-0), CNPq (407242/2021-0, 306573/2022-9, 442950/2023-3, 304711/2023-3), Labex Bézout/ANR, CAPES (88887.947543/2024-00, STIC-AMSUD 88887.878869/2023-00), and FAPEMIG (APQ-01079-23, PCE-00417-24 and APQ-05058-23).


\section*{materials}
The Parasites dataset is available at~\url{https://github.com/LIDS-Datasets/schistossoma-eggs}. The code to develop FLIM encoders and decoders will be available at~\url{https://github.com/LIDS-UNICAMP/flim-python-demo} after acceptance.

\bibliographystyle{IEEEtran}
\bibliography{refs.bib}

\end{document}